\PassOptionsToPackage{table}{xcolor}
\documentclass[runningheads]{llncs}

\usepackage[preprint]{colm} 

\setcitestyle{authoryear}

\usepackage{graphicx}
\usepackage{booktabs}
\usepackage[accsupp]{axessibility}
\usepackage{xltabular} 
\usepackage{microtype}
\usepackage{tabularx}
\usepackage{amssymb}
\usepackage{paralist}
\usepackage{todonotes}
\usepackage{breakurl}
\usepackage{subcaption}
\usepackage{latexsym}
\usepackage{amsmath}
\usepackage{float}
\usepackage{footnote}
\usepackage{enumitem}
\usepackage{bm}
\usepackage{arydshln}
\usepackage{multicol}
\usepackage{multirow}
\usepackage{colortbl}
\usepackage{bbding}
\usepackage{makecell}
\usepackage{mathtools}
\usepackage{imakeidx}
\usepackage{longtable}
\usepackage{tabularx}
\usepackage{wrapfig}
\usepackage{rotating}
\usepackage[edges]{forest}
\usepackage[normalem]{ulem}
\usepackage{CJKutf8}
\usepackage{awesomebox}
\usepackage[most]{tcolorbox}

\RequirePackage{xspace}
\makeatletter
\DeclareRobustCommand\onedot{\futurelet\@let@token\@onedot}
\def\@onedot{\ifx\@let@token.\else.\null\fi\xspace}

\def\eg{\emph{e.g}\onedot}

\makeatother

\usepackage{pifont}
\usepackage{lscape}
\usepackage{algorithm}
\usepackage{algpseudocode}

\usepackage{fancyhdr}
\renewcommand{\headrulewidth}{1pt}
\usepackage{fancyvrb}
\usepackage{fvextra}
\usepackage{amsfonts}
\usepackage{wrapfig}

\usepackage{hyperref}
\usepackage{orcidlink}

\definecolor{mydarkblue}{rgb}{0,0.08,0.45}
\definecolor{wkblue}{rgb}{0.2, 0.3, 0.6}
\definecolor{meta-color}{rgb}{0.5, 0.5, 0.5}
\definecolor{darkblue}{rgb}{0, 0, 0.5}
\definecolor{geovistagray}{gray}{0.95}
\definecolor{myblue}{rgb}{0.9, 0.1, 0.94}
\definecolor{mygreen}{rgb}{0.64, 0.56, 0.88}
\definecolor{myyellow}{rgb}{0.68, 0.6, 0.1}
\definecolor{fancygreen}{rgb}{0.33, 0.68, 0.20}
\definecolor{salmon}{rgb}{0.94, 0.52, 0.49}
\definecolor{tablegreen}{rgb}{0.82, 0.94, 0.75}
\definecolor{tableblue}{rgb}{0.81, 0.90, 0.94}
\definecolor{tablered}{rgb}{0.97, 0.85, 0.85}
\definecolor{tableorange}{rgb}{0.96, 0.85, 0.81}
\definecolor{bestcolor}{RGB}{210, 222, 239}
\definecolor{secondcolor}{RGB}{234, 239, 247}
\definecolor{thirdcolor}{RGB}{193, 214, 229}
\definecolor{line-blue}{RGB}{243, 248, 252}
\definecolor{line-green}{RGB}{200,242,200}
\definecolor{line-red}{RGB}{255,215,215}
\definecolor{line-gray}{RGB}{242, 242, 242}
\definecolor{sensepurple}{HTML}{5D2DD6}

\newenvironment{itemize*}%
 {\leftmargini=10pt\begin{itemize}%
  \setlength{\itemsep}{0pt}%
  \setlength{\parskip}{0pt}%
  }%
 {\end{itemize}}
\newenvironment{enumerate*}%
 {\begin{enumerate}%
  \setlength{\itemsep}{0pt}%
  \setlength{\parskip}{0pt}}%
 {\end{enumerate}}

\begin{document}

\title{Spatial-TTT: Streaming Visual-based Spatial Intelligence \\ with Test-Time Training}

\titlerunning{Spatial-TTT}

\author{\textbf{Fangfu Liu$^{*,1}$, Diankun Wu$^{*,1}$, Jiawei Chi$^{*,1}$,}
\textbf{Yimo Cai$^{1}$, Yi-Hsin Hung$^{1}$, Xumin Yu$^{2}$,}\\
\textbf{Hao Li$^{3}$, Han Hu$^{2}$, Yongming Rao$^{\dagger,2}$, Yueqi Duan$^{\dagger,1}$}\\[0.5em]
$^1$ Tsinghua University \quad $^2$ Tencent Hunyuan \quad $^3$ NTU}

\authorrunning{Liu et al.}

\maketitle

\makeatletter
\def\@makefnmark{}
\makeatother
\footnotetext{$^*$ Equal contribution. $^{\dagger}$ Corresponding author.}


\begin{center}
    \centering
    \vspace{-0.5cm}
    \includegraphics[width=1.0\linewidth]{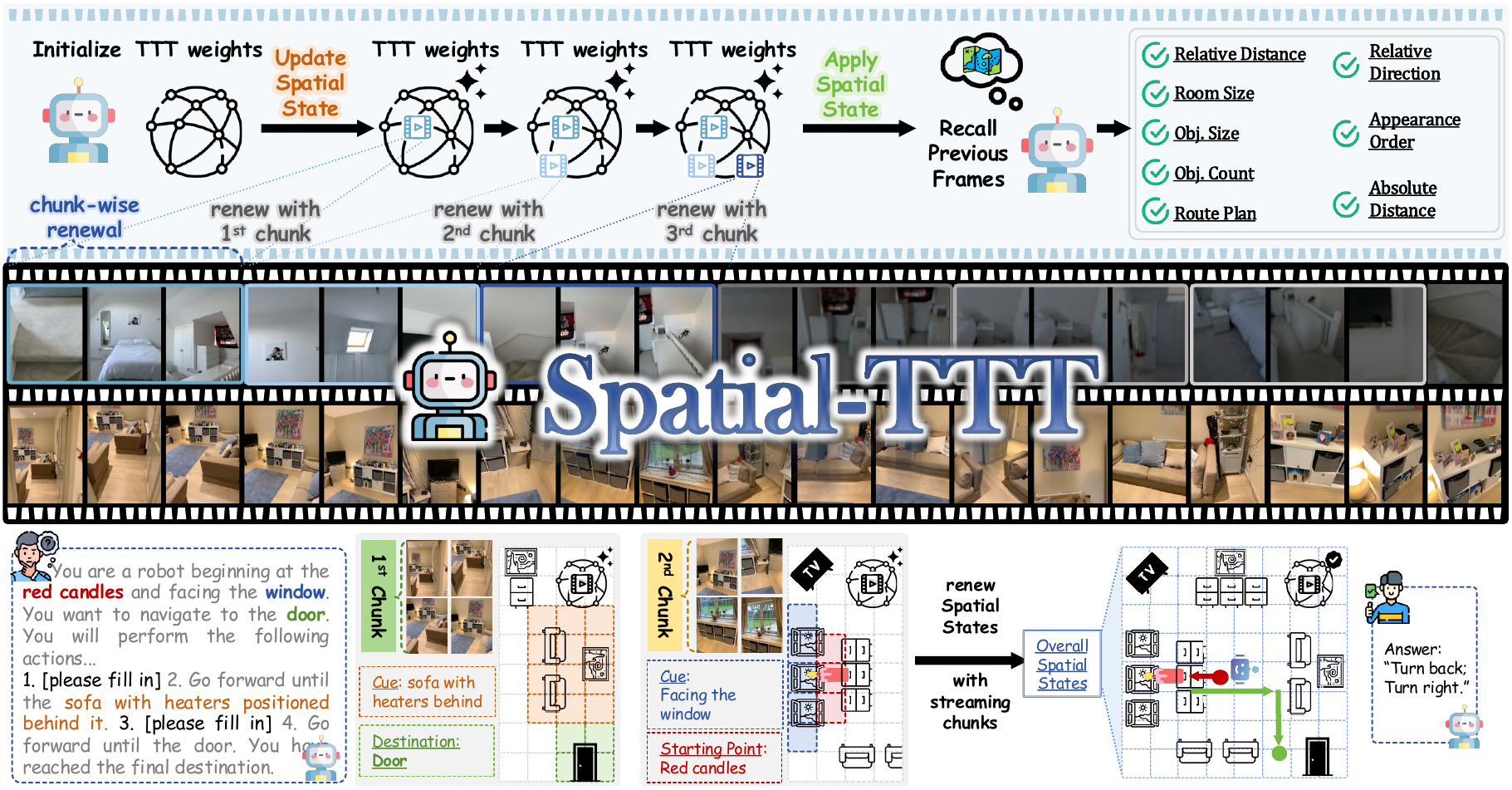}%
    \captionof{figure}
    {
    \textbf{Spatial-TTT}. Given a Visual-based Spatial task, our proposed Spatial-TTT updates spatial state with streaming chunks then answers the question.
    }
    \label{fig:teaser}
\end{center}%

\begin{abstract}
     Humans perceive and understand real-world spaces through a stream of visual observations. Therefore, the ability to streamingly maintain and update spatial evidence from potentially unbounded video streams is essential for spatial intelligence. The core challenge is not simply longer context windows but how spatial information is selected, organized, and retained over time. In this paper, we propose \textit{Spatial-TTT} towards streaming visual-based spatial intelligence with test-time training (TTT), which adapts a subset of parameters (fast weights) to capture and organize spatial evidence over long-horizon scene videos. Specifically, we design a hybrid architecture and adopt large-chunk updates parallel with sliding-window attention for efficient spatial video processing. To further promote spatial awareness, we introduce a spatial-predictive mechanism applied to TTT layers with 3D spatiotemporal convolution, which encourages the model to capture geometric correspondence and temporal continuity across frames. Beyond architecture design, we construct a dataset with dense 3D spatial descriptions, which guides the model to update its fast weights to memorize and organize global 3D spatial signals in a structured manner. Extensive experiments demonstrate that Spatial-TTT improves long-horizon spatial understanding and achieves state-of-the-art performance on video spatial benchmarks. Project page: \url{https://liuff19.github.io/Spatial-TTT}.

\end{abstract}

\section{Introduction}
Learning to perceive, understand, and reason about 3D structure and geometric relationships of the physical world is a cornerstone capability for spatial intelligence~\citep{fu2023mme, chen2024spatialvlm, yang2025cambriansspatialsupersensingvideo}, which finds broad applications in diverse fields including embodied robots~\citep{driess2023palme,huang2024rekep}, autonomous driving~\citep{hu2023uniad,wei2025spatial}, and augmented reality devices~\citep{grauman2022ego4d}. In these real-world scenarios, spatial information is rarely captured from a single static viewpoint but rather emerges from a continuous stream of visual observations, where the camera moves, objects become occluded, and viewpoints change over time~\citep{li2020streaming, yang2025cambriansspatialsupersensingvideo}. This necessitates streaming spatial understanding—the capability to selectively maintain, progressively update, and reason over spatial memory from long-horizon video inputs.

Recent advances in Multimodal Large Language Models (MLLMs) have demonstrated impressive results in 2D visual understanding~\citep{li2023blip, hurst2024gpt, zhu2025internvl3, bai2025qwen2, guo2025seed1}. However, their performance degrades significantly on tasks requiring spatial understanding, primarily due to the inherent lack of 3D geometric priors~\citep{Yang_2025_CVPR, yin2025spatial}, as these models are predominantly trained on 2D semantic-level image-text pairs without the supervision of spatial structure. While recent efforts have explored spatial-aware MLLMs through augmenting the input representations with geometric cues~\citep{hong20233dllm, wu2025spatialmllmboostingmllmcapabilities, fan2025vlm} with more 3D spatial VQA data~\citep{chen2024spatialvlm, yang2025visual}, they remain confined to single images or short video clips (\textit{i.e.,} 16 or 32 images) and cannot scale to the long-horizon video streams encountered in practical scenarios~\citep{yang2025cambriansspatialsupersensingvideo}, where spatial cues are scattered across thousands of frames and must be progressively aggregated as the observer navigates through the environment. Naively extending the input sequence leads to prohibitive computational costs due to quadratic attention complexity~\citep{vaswani2017attention, zhang2025ttt}, while aggressive temporal subsampling inevitably discards fine-grained spatial details critical for accurate 3D reasoning~\citep{team2024gemini, yang2025cambriansspatialsupersensingvideo}.

To address these challenges, we introduce \textit{Spatial-TTT}, a novel framework for streaming visual-based spatial intelligence built on the Test-Time Training (TTT) paradigm~\citep{sun2024learning}. Instead of using fixed parameters for inference, Spatial-TTT maintains adaptive fast weights that are updated online, acting as a compact non-linear memory to accumulate 3D evidence from unbounded video streams. To retain cross-modal alignment and semantic reasoning ability of pretrained MLLM, we employ a hybrid architecture that interleaves TTT layers with self-attention anchor layers, balancing efficient long-context compression with full-context reasoning. To make TTT practical for long spatial videos, we further adopt a large chunk update strategy~\citep{zhang2025ttt} for higher parallelism and hardware efficiency, and use sliding-window attention (SWA) in parallel to preserve intra-chunk spatiotemporal continuity. 

While these designs enable TTT scalable to long-horizon videos and are compatible with pretrained MLLMs, the Q/K/V used for updates are produced by point-wise linear projections, which ignore neighborhood structure among visual tokens and make the memory update dynamics less spatially coherent. To this end, we propose a \emph{spatial-predictive mechanism} that injects spatiotemporal inductive bias directly into the TTT branch. Instead of using point-wise linear projections, we apply lightweight depth-wise 3D spatiotemporal convolutions to aggregate local neighborhood context for visual tokens. This encourages the fast weights to learn predictive mappings between spatiotemporal contexts rather than isolated tokens, thereby better capturing geometric correspondence and temporal continuity and improving the stability and effectiveness of online updates.

Beyond architecture, effective TTT requires supervision that teaches the model how to update fast weights to preserve globally useful 3D evidence over long streams. However, existing spatial datasets are sparse and local, providing weak gradient signals for model to learn fast-weight update dynamics. To address this, we construct a dense scene-description dataset where the model is required to generate comprehensive 3D scene descriptions covering global context, objects and counts, and spatial relations. This dense description provides rich supervision for training fast-weight update dynamics to preserve structured, scene-level spatial information along video stream.
Extensive experiments demonstrate that Spatial-TTT significantly improves long-horizon spatial understanding and achieves state-of-the-art performance on video spatial benchmarks. Our main contributions are summarized as follows: 
\begin{itemize}[leftmargin=*]
    \item We propose {Spatial-TTT} for streaming visual-based spatial intelligence with test-time training, which performs online fast-weight updates as compact non-linear memory to accumulate spatial evidence from long-horizon spatial video streams.
    \item We design a hybrid test-time training architecture together with large-chunk updates and parallel sliding-window attention, achieving efficient long spatial-context compression and reasoning.
    \item We introduce a spatial-predictive mechanism that captures geometric correspondence and spatial-temporal continuity, and further construct a dense scene-description dataset to provide rich supervision for learning effective fast-weight update dynamics.
    \item We conduct extensive experiments on video spatial benchmarks, which demonstrate that our method achieves state-of-the-art performance on a wide range of visual-based spatial understanding tasks.
\end{itemize}

\section{Related Work}

\subsection{Visual-based Spatial Intelligence}
Visual-based Spatial Intelligence of MLLMs focuses on MLLMs’ ability to perceive, reason about, and recall spatial relationships as well as layouts from visual inputs. While most existing MLLMs demonstrate strong performance on 2D perception and reasoning tasks \citep{liu2024llava, Qwen3-VL}, they still struggle with tasks that require precise 3D spatial alignment, such as robotic manipulation \citep{li2024manipllm} and 3D question answering \citep{hong20233dllm, xu2023pointllm, chen2024spatialvlm}.  To minimize this gap, previous researchers have constructed several benchmarks~\citep{li2024mvbench, Yang_2025_CVPR, yin2025spatial, yang2025cambriansspatialsupersensingvideo}, for example, VSI-Bench for evaluating comprehensive video-based visual-spatial intelligence \citep{Yang_2025_CVPR}, STI-Bench for examining spatial-temporal understanding \citep{Li_2025_ICCV}, and VSI-Super for challenging spatial recall and continual counting \citep{yang2025cambriansspatialsupersensingvideo}. Meanwhile, several methods were proposed to enhance visual-based spatial intelligence of MLLMs, like incorporating metric
depth and multi-view inputs in MM-Spatial \citep{Daxberger_2025_ICCV}, adopting feed-forward visual geometry models in Spatial-MLLM \citep{wu2025spatialmllmboostingmllmcapabilities} and VLM-3R~\citep{fan2025vlm}, SFT and RL methods explored by SpaceR \citep{ouyang2025spacer} and MindCube \citep{yin2025spatial}, and 3D feature alignment at output proposed by 3DThinker \citep{chen2025think}. Recently, VST \citep{yang2025visual} constructed 4.1M SFT dataset for spatial perception and 135K RL dataset for spatial reasoning. SpatialLadder \citep{li2025spatialladder} built a 26K dataset, and Cambrian-S \citep{yang2025cambriansspatialsupersensingvideo} proposed a four-stage training framework with VSI-590K dataset. However, existing methods mostly focused on pre-training or post-training stages while test-time strategy for natively adapting diverse and streaming data is still not fully explored.

\subsection{Test-Time Training}
Test-time training (TTT) refers to methods that improve model performance using only unlabeled test data at inference time \citep{ba2016using, sun2020test, wang2021tent, gandelsman2022test}. Unlike test-time scaling (TTS), which relies on sampling multiple trajectories and selecting the most promising ones with frozen model parameters at test time \citep{snell2024scaling, deepseek2025r1, openai2024o1}, test-time training continually updates model parameters during inference to adapt to diverse inputs and tasks \citep{schlag2021linear, Liang_2024}. Prior work has shown that adapting fast weights over large chunks at inference time enables efficient memorization, with applications spanning novel view synthesis, language modeling, and autoregressive video diffusion \citep{zhang2025ttt}. These results highlight the potential of TTT for long-context visual tasks, such as spatial intelligence in streaming video settings \citep{zhang2025ttt}. More recent studies, including TTT-E2E \citep{tandon2025endtoend} and work on TTT for few-shot learning \citep{akyurek2025surprising}, demonstrate that TTT supports continual weight adaptation during inference in an end-to-end manner and yields substantial reasoning improvements beyond in-context learning. Researchers have also developed broader design space, including optimizers, various loss functions and neural representation of memory \citep{wang2025test, behrouz2024titans, karami2025lattice}. While TTT has achieved notable success in language modeling, its application to enhancing the visual capabilities of MLLMs has received comparatively limited attention \citep{shu2022tpt, sun2024learningtolearn}.
\begin{figure*}[t]
    \centering
    \includegraphics[width=\textwidth]{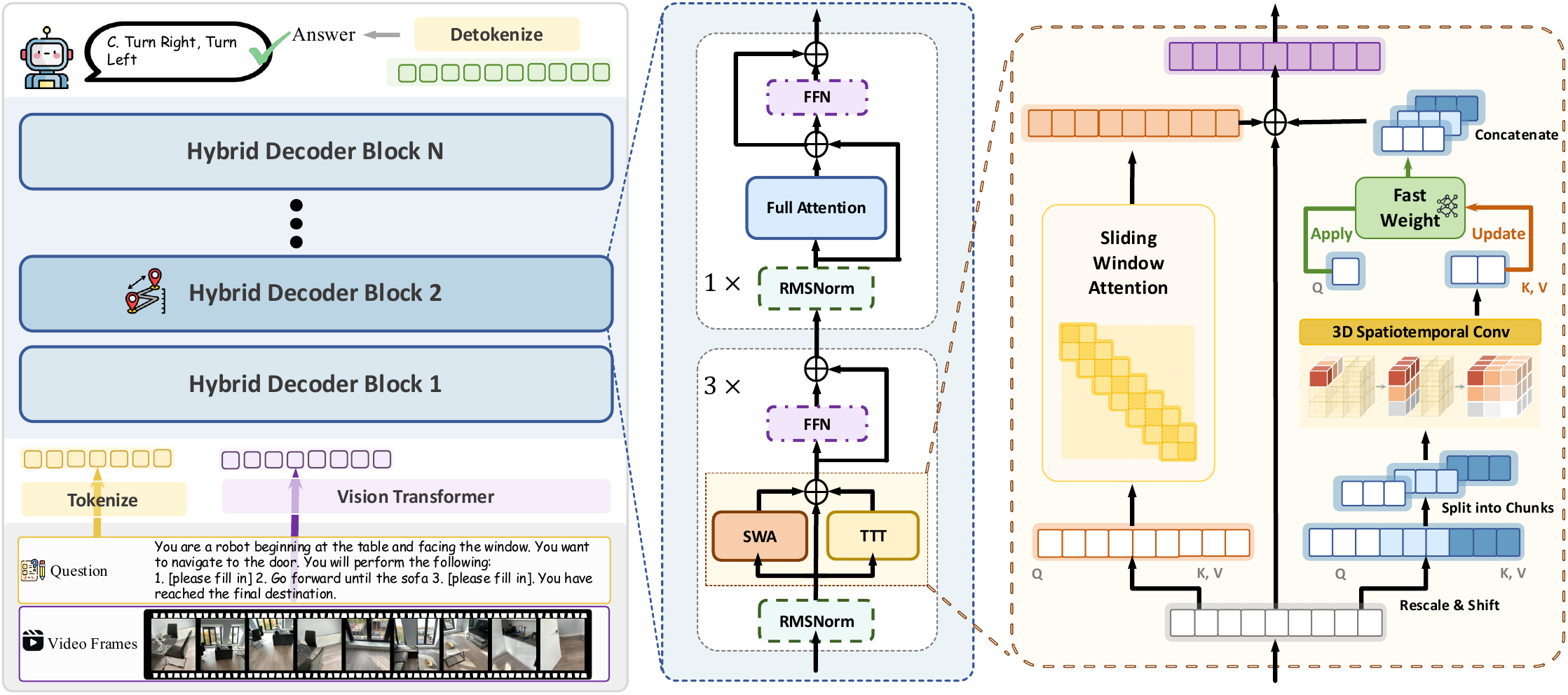}
    \caption{\textbf{Overview of Spatial-TTT}. The model employs a hybrid architecture that interleaves TTT layers with self-attention anchor layers at a 3:1 ratio to preserve pretrained knowledge while enabling efficient long spatial-context compression. Within each TTT layer, sliding window attention (SWA) and TTT branch operate in parallel with shared Q/K/V projections, where the TTT branch applies spatial predictive mechanism with depthwise spatiotemporal convolution to capture geometric structure and temporal continuity.}
    \label{fig:pipeline}
\end{figure*}

\section{Spatial-TTT}
In this section, we first introduce the overall framework (Sec.~\ref{subsec:overall-framework}) of Spatial-TTT shown in Fig.~\ref{fig:pipeline}, which consists of a hybrid TTT architecture that interleaves TTT layers with standard self-attention layers to preserve pretrained visual-semantic knowledge, and a spatial-predictive mechanism that enhances TTT with lightweight depthwise convolutions to learn spatiotemporal structure. We then describe how to bridge sparse spatial supervision with dense scene descriptions that provide rich, scene-level gradient signals for learning effective fast-weight update dynamics (Sec.~\ref{subsec:spatialqa}). Finally, we present a spatial-aware progressive training strategy (Sec.~\ref{subsec:training}) towards effective visual-based spatial reasoning capability. Before introducing our method in detail, we first review the theory of Test-Time Training (TTT).

\subsection{Preliminary}
\textbf{Test-Time Training.} Traditional neural networks follow a static paradigm where model parameters $\theta$ remain frozen after training, limiting their ability to dynamically adapt to evolving inputs during inference. Test-Time Training (TTT) offers an alternative paradigm that enables on-the-fly parameter adaptation by updating a designated subset of parameters (\textit{i.e.,} fast weights) while processing each test sequence. The TTT mechanism maintains fast weights $W \in \mathbb{R}^{d_{out} \times d_{in}}$ that parameterize a small neural network $f_W: \mathbb{R}^{d_{in}} \rightarrow \mathbb{R}^{d_{out}}$. Unlike conventional model parameters that are fixed post-training, these fast weights function as an adaptive memory that continuously encodes contextual information from the input stream. Given an input sequence $\mathbf{x} = [x_1, x_2, \ldots, x_T]$ where $x_t \in \mathbb{R}^d$, each token is projected to derive a key $k_t$, a query $q_t$, and a value $v_t$. TTT alternates between two operations at each timestep:
\begin{enumerate}[leftmargin=*]
    \item \textbf{Update Operation:} The fast weights are modified to associate the current key-value pair $(k_t, v_t)$ by performing a gradient descent step on a self-supervised loss function $\mathcal{L}$:
    \begin{equation}
        W_t \leftarrow W_{t-1} - \eta \nabla_W \mathcal{L}\big(f_{W_{t-1}}(k_t), v_t\big).
        \label{eq1}
    \end{equation}
    where $\eta$ denotes the learning rate. This step encodes information from the pair $(k_t, v_t)$ into the neural memory $W_t$.
    \item \textbf{Apply Operation:} The updated network produces the output by processing the query:
    \begin{equation}
        o_t = f_{W_t}(q_t).
    \end{equation}
    The output $o_t$ is enriched with contextual information from all preceding tokens, as their key-value associations have been progressively encoded into $W_t$.
\end{enumerate}

\subsection{Overall Framework}
\label{subsec:overall-framework}

\textbf{Hybrid TTT Architecture.}
The inherent linear complexity of Test-Time Training makes it well-suited for processing unbounded spatial streams from long-horizon videos. However, directly replacing all core attention layers with TTT layers is prone to disrupting the pretrained cross-modal alignment and visual semantics, while recovering these would require heavily retraining~\citep{sun2020test, sun2024learning}. To address this, we employ a \textit{hybrid TTT architecture} that interleaves TTT layers with self-attention layers at a 3:1 ratio. In a transformer with $L$ decoder layers, 75\% use TTT while the remaining 25\% retain standard self-attention as {anchor layers}. The anchor layers maintain full attention access over the entire context, preserving the pretrained model's semantic reasoning ability. Meanwhile, TTT layers compress long-range temporal dependencies into adaptive fast weights $W_t$, achieving sublinear memory growth.

Within each TTT layer, conventional implementations use small chunks (\textit{e.g.}, 16 or 64 tokens) for frequent fast-weight updates~\citep{sun2023retentive}. This works poorly for streaming visual observations: small chunks lead to low GPU utilization due to poor parallelism, and chunk boundaries artificially break the spatial structure of video frames. Inspired by LaCT~\citep{zhang2025ttt}, we instead adopt a large chunk size for visual tokens, roughly aligned with multiple video frames, to substantially improve parallelism and hardware efficiency and to keep spatially coherent visual content inside the same update unit.
A key challenge with large chunks arises from the causal constraints in TTT updates: to prevent information leakage, the current chunk cannot interact with itself during fast-weight updates, which would otherwise allow earlier tokens to access later ones. This restriction precludes any intra-chunk token interactions, which is undesirable for maintaining spatiotemporal continuity in visual data. To address this, we incorporate sliding window attention (SWA) within each TTT layer, operating in parallel with TTT, sharing query, key, and value projections (with lightweight learnable scale and shift for TTT's queries and keys). The window size $w$ is set to be no smaller than the chunk size $b$, so that the SWA fully covers the causal lower-triangular attention matrix within each chunk, ensuring the completeness of the causal structure. The layer output combines both branches:

\begin{equation}
    o_t = {\mathrm{WindowAttn}(q_t, K_{[t-w:t]}, V_{[t-w:t]})} + {f_{W_t}(q_t)},
\end{equation}
where $K_{[t-w:t]}$ and $V_{[t-w:t]}$ denote the keys and values within the sliding window.
For the fast-weight network $f_W$, we use a bias-free SwiGLU-MLP to increase the nonlinearity and expressiveness of the memory:
\begin{equation}
    f_W(\mathbf{x}) = W_2 \big[ \mathrm{SiLU}(W_1 \mathbf{x}) \odot (W_3 \mathbf{x}) \big],
\end{equation}
where $\mathbf{x}$ denotes the input ($k_t$ for update, $q_t$ for apply), $W = \{W_1, W_2, W_3\}$ are the learnable fast weights, and $\odot$ is element-wise multiplication.

\noindent \textbf{Spatial-Predictive Mechanism.} Streaming spatial understanding poses unique challenges: spatial information emerges from continuous visual observations, where the camera moves, viewpoints change, and objects are gradually revealed or occluded. Adjacent visual tokens describe a progressively unfolding 3D scene with strong geometric and temporal continuity. However, in conventional TTT designs~\citep{sun2020test, zhang2025ttt}, Q, K, V are generated through point-wise linear projections: $k_t = W_k x_t$, $v_t = W_v x_t$, which ignore spatial-temporal structure among visual tokens. Since fast weights must compress the chunk into compact memory to represent 3D structure, using isolated token transforms for K, V without local geometric context forces them to learn spatial patterns from scratch—making compression harder and memory less structured. To address this, we introduce a spatial-predictive mechanism with lightweight depth-wise spatiotemporal convolution on the Q, K, V of the TTT branch. For visual tokens from videos, we reshape them into a spatiotemporal grid and aggregate neighborhood information through local aggregation. Formally, for the $i$-th channel of a token at spatiotemporal position $(t, h, w)$, the conv-enhanced query, key and value are:
\begin{equation}
    \tilde{x}_{t,h,w}^{\,i} = \sum_{\delta \in \mathcal{N}} \theta_\delta^{\,i} \cdot x_{t+\delta_t, h+\delta_h, w+\delta_w}^{\,i}, \quad x \in \{q, k, v\},
\end{equation}
where $\mathcal{N} = \{-\lfloor\kappa/2\rfloor, ..., \lfloor\kappa/2\rfloor\}^3$ is the local neighborhood with kernel size $\kappa$, and $\theta \in \mathbb{R}^{\kappa^3 \times d}$ are learnable kernel weights initialized with Dirac delta to preserve identity mapping at initialization. To further improve the stability and effectiveness of the TTT update, we adopt the Muon update rule~\citep{jordan2024muon, zhang2025ttt} instead of vanilla implementation (\textit{i.e.,} Equation~\ref{eq1}) for optimization:
\begin{align}
    G_t &= \text{MuonUpdate}(G_{t-1}, \nabla_W \mathcal{L}(f_{W_{t-1}}(\tilde{k}_t), \tilde{v}_t)), \\
    W_t &\leftarrow \text{L2Norm}(W_{t-1} - \eta G_t),
\end{align}
where $G_t$ is the orthogonalized gradient with momentum, $\text{Muon}(\cdot, \cdot)$ accumulates momentum and orthogonalizes via Newton-Schulz iterations, and L2Norm normalizes weights while preserving their original magnitude. Powered by the spatial-predictive mechanism, fast weights no longer learn prediction between isolated tokens, but rather predictive mapping between spatial-temporal contexts. This enables fast weights to implicitly capture geometric correspondence and temporal continuity in streaming spatial understanding.

\subsection{Bridging Sparse Spatial QA with Dense Supervision}
\label{subsec:spatialqa}
The effectiveness of TTT~\citep{sun2024learning} depends on whether the model learns fast-weight update dynamics that retain information useful for future timesteps. However, supervision in existing spatial intelligence datasets~\citep{wu2025spatialmllmboostingmllmcapabilities, yang2025cambriansspatialsupersensingvideo, fan2025vlm} is typically sparse and local. For instance, a typical spatial QA task queries relations between two objects in a small region of the scene (often answerable from only a few frames), and the target answers are often short (\textit{e.g.,} a multiple-choice option or an integer). Such target answers only supervise a small subset of the underlying 3D scene states, providing weak and low-coverage gradient signals for learning fast-weight update dynamics. Consequently, the model has limited incentive to construct a coherent and persistent global 3D memory over long video streams.

To bridge this gap, we construct a dense scene-description dataset from SceneVerse annotations~\citep{jia2024sceneverse}. For each training example, the model receives a spatial video stream and is required to generate a comprehensive description of the underlying 3D scene. Unlike short-form answers, the target output is formatted as a coherent scene walkthrough, which is derived from object-centric 3D scene graphs~\citep{jia2024sceneverse}. Specifically, the target description include following aspects: (1) \emph{Global context:} identifying the scene type and functional setting, encouraging fast weights to encode global semantic descriptors beyond local visual cues; (2) \emph{Objects and counts:} enumerating object categories and precise counts, which encourages retention of persistent instance-level evidence across time; and (3) \emph{Object relations:} describing spatial layouts and pairwise relations, which promotes encoding of geometric structure and inter-object constraints.
This dense scene description task provides rich and high-coverage supervision that complements sparse spatial QA. By training the model to generate detailed 3D scene descriptions, we encourage the fast-weight dynamics to capture global and persistent spatial representations that benefit downstream spatial reasoning.

\subsection{Spatial-Aware Progressive Training}
\label{subsec:training}
With the architecture and data designed above, the remaining question is how to train the model effectively. We meticulously design a two-stage spatial-aware progressive training strategy: (1) initializing fast weights with global 3D awareness using dense scene description data, and (2) tuning streaming spatial reasoning capability with large-scale spatial VQA data. Specifically, we first train the hybrid TTT architecture on the dense scene description dataset described above, with the goal of teaching fast weights to retain comprehensive scene-level information through chunk-by-chunk memory updates. To better preserve the pretrained visual knowledge during this process, we do not directly set the sliding window size equal to the chunk size. Instead, we employ a sliding window annealing strategy: the sliding window size $w$ is linearly annealed from an initial value $w_{\max}$ to $w_{\min} = b$ (the chunk size) over the first stage. As the window size gradually decreases, TTT layers are forced to take over more responsibility for cross-chunk information propagation while fast weights progressively learn to encode global 3D scene structure during memory updates. However, spatial understanding requires not only ``remembering" spatial information but also effectively ``recalling and reasoning" about spatial relations during streaming observations. To further promote streaming visual-based spatial intelligence, we fine-tune the model in the second stage using 2M spatial VQA samples covering diverse tasks such as object relative direction/distance estimation, spatial counting, route planning, and room size estimation. In this stage, the window size and chunk size are fixed at the same value ($w = b$), so TTT layers are fully focused on cross-chunk spatial information aggregation. Through this fine-tuning, the model learns to selectively retain task-relevant spatial evidence via fast weight updates and retrieve accumulated spatial knowledge when reasoning.

At inference time, we employ a dual KV cache mechanism for constant-memory streaming. The first is a sliding window KV cache of fixed length $w$ for local context modeling in sliding window attention; when the cache exceeds the window size, the earliest entries are discarded. The second is a TTT pending KV cache that accumulates key-value pairs for fast weight updates: it starts empty and grows as new tokens arrive; whenever its length reaches the chunk size $b$, these KV pairs are used to perform one fast weight update, after which the pending cache is cleared.

\section{Experiments}
\begin{table*}[t]
\centering
\caption{\textbf{Evaluation Results on VSI-Bench~\citep{Yang_2025_CVPR}.}
For Numerical Questions, we report MRA score. For Multiple-Choice Questions, we report ACC score, Avg. is the macro average across all tasks, following the original paper. For human, we directly use the reported results in VSI-Bench, which is 
on a subset of VSI-Bench with 400 samples. We use \colorbox{bestcolor}{\hspace{1.2em}} and \colorbox{secondcolor}{\hspace{1.2em}} to denote the best and second-best results within proprietary models and open-source models, respectively.}
\label{tab:vsibench}

\renewcommand{\arraystretch}{1.1} 
\setlength{\tabcolsep}{1pt} 

\resizebox{1.0\linewidth}{!}{
\begin{tabular}{l*{9}{c}}
\toprule
\multirow{2}{*}{\textbf{Models}} &
\multicolumn{4}{c}{\textbf{Numerical Question}} &
\multicolumn{4}{c}{\textbf{Multiple-Choice Question}} &
\multirow{2}{*}{\textbf{Avg.}} \\
\cmidrule(lr){2-5}\cmidrule(lr){6-9}
&
\makecell{\scriptsize \textbf{Obj. Count}} &
\makecell{\scriptsize \textbf{Abs. Dist}} &
\makecell{\scriptsize \textbf{Obj. Size}} &
\makecell{\scriptsize \textbf{Room Size}} &
\makecell{\scriptsize \textbf{Rel. Dis}} &
\makecell{\scriptsize \textbf{Rel. Dir}} &
\makecell{\scriptsize \textbf{Route Plan}} &
\makecell{\scriptsize \textbf{Appr. Order}} &
\multicolumn{1}{c}{} \\

\midrule
Human & 94.3 & 47.0 & 60.4 & 45.9 & 94.7 & 95.8 & 95.8 & 100.0 & 79.2 \\

Random Choice & 62.1 & 32.0 & 29.9 & 33.1 & 25.1 & 47.9 & 28.4 & 25.2 & 34.0 \\

\midrule
\rowcolor{line-gray}\multicolumn{10}{l}{\textbf{Proprietary Models}} \\

    


Seed-2.0~\citep{seed20} & 49.4 & 25.3 & 69.5 & 25.8 & 61.8 & 44.9 & 44.3 & 71.0 & 50.7 \\

Grok-4~\citep{grok4_xai_2025}
& 37.1 & 32.9 & 60.8 & 45.4 & 53.1 & 39.6 & 47.4 & 66.8 & 47.9 \\

Gemini-2.5-pro~\citep{comanici2025gemini}
& 46.0 & \cellcolor{secondcolor}37.3 & 68.7 & \cellcolor{secondcolor}54.3 & \cellcolor{secondcolor}61.9 & 43.9 & 47.4 & \cellcolor{secondcolor}{68.7} & 53.5 \\

Gemini-3-pro~\citep{gemini3} & 49.0 & \cellcolor{bestcolor}\textbf{42.8} & \cellcolor{secondcolor}71.5 & 41.8 & 56.6 & \cellcolor{bestcolor}\textbf{57.5} & \cellcolor{bestcolor}\textbf{61.9} & 60.0 & \cellcolor{bestcolor}\textbf{56.0} \\

Kimi-K2.5~\citep{kimiteam2026kimik25visualagentic} & \cellcolor{bestcolor}\textbf{57.2} & 34.9 & 69.3 & \cellcolor{bestcolor}\textbf{54.4} & 59.6 & 41.3 & \cellcolor{secondcolor}52.1 & 67.0 & 53.6 \\

GPT-5~\citep{openai_gpt5_systemcard}
& \cellcolor{secondcolor}{53.3} & 34.4 & \cellcolor{bestcolor}{\textbf{73.3}} & 47.5 & \cellcolor{bestcolor}{\textbf{63.7}} & \cellcolor{secondcolor}48.6 & 50.2 & \cellcolor{bestcolor}{\textbf{68.9}} & \cellcolor{secondcolor}55.0 \\

\midrule
\rowcolor{line-gray}\multicolumn{10}{l}{\textbf{Open-source General Models}} \\

LongVA-7B~\citep{zhang2024long} & 38.0 & 16.6 & 38.9 & 22.2 & 33.1 & 43.3 & 25.4 & 15.7 & 29.2\\
LLaVA-OneVision-72B~\citep{li2024llava}  & 43.5 & 23.9 & 57.6 & 37.5 & 42.5 & 39.9 & 32.5 & 44.6 & 40.2\\

LLaVA-Video-72B~\citep{lin2023videollava}  & 48.9 & 22.8 & 57.4 & 35.3 & 42.4 & 36.7 & 35.0 & 48.6 & 40.9\\

InternVL3-2B~\citep{zhu2025internvl3} 
& 64.8 & 30.8 & 32.4 & 22.9 & 32.2 & 34.9 & 32.9 & 12.6 & 32.9 \\

InternVL3-8B~\citep{zhu2025internvl3}
& 66.0 & 34.8 & 43.6 & 47.5 & 48.0 & 39.3 & 26.2 & 31.3 & 42.1 \\

Qwen2.5-VL-3B-Instruct~\citep{bai2025qwen2}   & 24.3 & 24.7 & 31.7 & 22.6 & 38.3 & 42.6 & 26.3 & 21.2 & 29.0\\

Qwen2.5-VL-7B-Instruct~\citep{bai2025qwen2}  & 40.9 & 14.8 & 43.4 & 10.7 & 38.6 & 40.1 & 33.0 & 29.8 & 31.4\\

Qwen3-VL-2B-Instruct~\citep{Qwen3-VL}
& 62.1 & 40.2 & 71.4 & 49.7 & 52.2 & 42.0 & 30.4 & 54.5 & 50.3 \\

Qwen3-VL-8B-Instruct~\citep{Qwen3-VL}
& 67.5 & \cellcolor{secondcolor}{47.0} & \cellcolor{bestcolor}{\textbf{76.3}} & 61.9 & 58.0 & 50.9 & 35.0 & 66.3 & 57.9 \\

\midrule
\rowcolor{line-gray}\multicolumn{10}{l}{\textbf{Open-source Spatial Intelligence Models}} \\

MindCube-3B~\citep{yin2025spatial}
& 12.8 & 22.7 & 4.3 & 23.4 & 20.2 & 15.7 & 15.9 & 22.4 & 17.2 \\

SpatialLadder-3B~\citep{li2025spatialladder}
& 62.1 & 35.3 & 61.9 & 41.4 & 45.6 & 46.4 & 27.3 & 38.5 & 44.8 \\

SpaceR-7B~\citep{ouyang2025spacer} 
& 44.5 & 24.7 & 53.5 & 37.3 & 41.9 & 46.1 & 29.3 & 54.8 & 41.5 \\

ViLaSR-7B~\citep{wu2025reinforcing}
& 58.1 & 33.8 & 61.4 & 28.8 & 45.0 & 46.5 & 29.9 & 53.2 & 44.6 \\

VST-3B-SFT~\citep{yang2025visual}
& 69.3 & 45.4 & 71.8 & 62.4 & 59.0 & 46.0 & 38.7 & 70.2 & 57.9 \\

VST-7B-SFT~\citep{yang2025visual}
& \cellcolor{bestcolor}{\textbf{72.0}} & 44.4 & \cellcolor{secondcolor}{74.3} & \cellcolor{bestcolor}{\textbf{68.3}} & 59.7 & 55.8 & \cellcolor{secondcolor}{44.9} & 65.2 & \cellcolor{secondcolor}{60.6} \\

Cambrian-S-3B~\citep{yang2025cambriansspatialsupersensingvideo}
& 70.7 & 40.6 & 68.0 & 46.3 & \cellcolor{bestcolor}{\textbf{64.8}} & \cellcolor{secondcolor}{61.9} & 27.3 & \cellcolor{bestcolor}{\textbf{78.8}} & 57.3 \\

Spatial-MLLM-4B~\citep{wu2025spatialmllmboostingmllmcapabilities}  & 65.3 & 34.8 & 63.1 & 45.1 & 41.3 & 46.9 & 33.5 & 46.3 & 47.0\\

\midrule
\rowcolor{line-gray}\multicolumn{10}{l}{\textbf{Ours}} \\

\textbf{Spatial-TTT-2B}  & \cellcolor{secondcolor}{70.8} & \cellcolor{bestcolor}{\textbf{47.8}} & 71.7 & \cellcolor{secondcolor}{65.9} & \cellcolor{secondcolor}{61.8} & \cellcolor{bestcolor}{\textbf{73.0}} & \cellcolor{bestcolor}{\textbf{47.4}} & \cellcolor{secondcolor}{77.0} & \cellcolor{bestcolor}{\textbf{64.4}} \\

\bottomrule
\end{tabular}
}

\end{table*}

\subsection{Experiment Setup}

\textbf{Implementation Details.}
Our model is initialized from Qwen3-VL-2B-Instruct~\citep{Qwen3-VL}. In the hybrid TTT architecture design, we apply TTT layers to every three of every four attention layers. To preserve pretrained knowledge and facilitate convergence, we do not initialize new QKV projection matrices from scratch and make the TTT layers share the original attention layer's QKV projection matrix. To increase expressivity of TTT layers, we introduce lightweight learnable scale-and-shift parameters applied to the projected $q$ and $k$. At the start of training, the gating projection is initialized to zero, the scale parameters are set to ones, and the shift parameters are set to zeros. In the spatial-predictive mechanism, we apply depthwise 3D convolutions with kernel size $3 \times 3 \times 3$ on the Q, K, V projections of TTT layers, which is Dirac-initialized, ensuring that the network initially behaves identically to the original full-attention model. The LaCT chunk size $b$ is set to 2648, and the window size $w$ is initialized to 5600—sufficient to cover the entire sequence—and annealed to 2648 over the first two epochs on the dense scene-description dataset, where 32 frames are uniformly sampled for training. We then perform continuous finetuning with large-scale spatial VQA data from 64 to 128 frames. We use a learning rate of 1e-6 for the pretrained backbone and 1e-5 for the newly introduced TTT-related parameters, with a warmup of 1k steps and a cosine learning-rate scheduler across all training stages.

\noindent \textbf{Datasets.}
In the first stage, we train our model on the dense scene-description dataset (Sec.~\ref{subsec:spatialqa}). This dataset contains approximately 16k samples, including 3.6k descriptions of ScanNet~\citep{dai2017scannet} scenes and 12.5k descriptions of ARKitScenes~\citep{baruch2021arkitscenes} scenes. In the second stage, we train the model on a 3M large-scale spatial question-answering dataset, which includes open-sourced spatial data and self-constructed data. Detailed statistics and curation procedures are provided in the Supplementary Materials.

\noindent \textbf{Baselines.}
We compare our model against a broad set of baselines, covering proprietary multimodal models (including GPT-5~\citep{openai_gpt5_systemcard}, Gemini-3-Pro~\citep{gemini3}, Seed-2.0~\citep{seed20}, Kimi-K2.5~\citep{kimiteam2026kimik25visualagentic} etc.), open-source general-purpose MLLMs (LLaVA-OneVision~\citep{li2024llava}, LLaVA-Video~\citep{lin2023videollava}, InternVL3 series~\citep{zhu2025internvl3}, Qwen2.5-VL series~\citep{bai2025qwen2}, Qwen3-VL series~\citep{Qwen3-VL}, etc.), and open-source spatial-intelligence models (VST~\citep{yang2025visual}, Cambrian-S~\citep{yang2025cambriansspatialsupersensingvideo}, Spatial-MLLM~\citep{wu2025spatialmllmboostingmllmcapabilities}, etc.). For long-horizon recall and counting on VSI-SUPER~\citep{yang2025cambriansspatialsupersensingvideo}, we additionally include long-video understanding baselines (MovieChat~\citep{song2024moviechat} and Flash-VStream~\citep{zhang2024flash}). Additionally, we report Human and Random Choice scores for VSI-Bench~\citep{Yang_2025_CVPR} and MindCube~\citep{yin2025spatial} as references.

\subsection{General Spatial Understanding Results}
\textbf{Results on VSI-Bench.}
To assess the models’ general spatial understanding capabilities, we evaluate our model and the baseline models on VSI-Bench~\citep{Yang_2025_CVPR}, a benchmark that measures multimodal visual-spatial intelligence through egocentric video understanding. VSI-Bench contains over 5,000 question–answer pairs constructed from 288 real-world indoor videos drawn from the validation splits of ScanNet~\citep{dai2017scannet}, ScanNet++~\citep{yeshwanth2023scannet++}, and ARKitScenes~\citep{baruch2021arkitscenes}, covering diverse environments (\textit{e.g.,} homes, offices, labs, and factories) across multiple geographic regions. Following the original paper~\citep{Yang_2025_CVPR}, we use Accuracy (ACC) for multiple-choice questions and Mean Relative Accuracy (MRA) for numerical questions to quantify how closely predictions align with ground-truth values. Avg. is the macro average
across these metrics. The results are reported in Table~\ref{tab:vsibench}.
As shown, our Spatial-TTT-2B achieves the best overall performance on VSI-Bench with an Avg. of 64.4, surpassing both proprietary and open-source baselines despite its compact 2B scale. In particular, our model exhibits strong advantages on multiple-choice spatial reasoning tasks that require consistent geometric understanding across egocentric views, including Relative Direction and Route Plan, indicating improved capability in direction reasoning and navigation planning. Meanwhile, on numerical questions, our model attains the best score on Absolute Distance and remains highly competitive on Room Size and Object Count, suggesting stronger metric grounding and scene-scale estimation than other baselines.

\begin{wraptable}{r}{7.6cm} 
    \centering
    \vspace{-0.8cm}
    \renewcommand{\arraystretch}{1.2} 
    \caption{\textbf{Evaluation results on MindCube-Tiny~\citep{yin2025spatial}.} Avg. is micro average across all tasks, following the original paper. We use \colorbox{bestcolor}{\hspace{1.2em}} and \colorbox{secondcolor}{\hspace{1.2em}} to denote the best and second-best results within proprietary models and open-source models, respectively.}
    \label{tab:mindcube}
    
    \scriptsize 
    \setlength{\tabcolsep}{1pt} 
    
    \resizebox{\linewidth}{!}{%
        \begin{tabular}{lcccc}
            \toprule
            \textbf{Models} & \textbf{Avg.} & \textsc{Rotation} & \textsc{Among} & \textsc{Around} \\
            
            \midrule
            Human & 94.5 & - & - & - \\
            Random Choice & 33.0 & 33.3 & 31.8 & 35.7 \\
            
            \midrule
            \rowcolor{line-gray}\multicolumn{5}{l}{\textbf{Proprietary Models}} \\
            Seed-2.0~\citep{seed20} & 54.8 & 89.0 & 45.2 & 52.0 \\
            Grok-4~\citep{grok4_xai_2025} & \cellcolor{secondcolor}63.5 & \cellcolor{secondcolor}{93.0} & \cellcolor{secondcolor}54.4 & 61.6 \\
            Gemini-2.5-pro~\citep{comanici2025gemini} & 57.6 & 88.0 & 44.9 & 63.2 \\
            Gemini-3-pro~\citep{gemini3} & \cellcolor{bestcolor}\textbf{63.9} & 73.5 & \cellcolor{bestcolor}\textbf{59.3} & \cellcolor{secondcolor}{66.0} \\
            Kimi-K2.5~\citep{kimiteam2026kimik25visualagentic} & 57.7 & 82.5 & 50.2 & 56.5 \\
            GPT-5~\citep{openai_gpt5_systemcard} & 56.3 & \cellcolor{bestcolor}{\textbf{94.5}} & 38.2 & \cellcolor{bestcolor}{\textbf{68.4}} \\
            
            \midrule
            \rowcolor{line-gray}\multicolumn{5}{l}{\textbf{Open-source General Models}} \\
            InternVL3-2B~\citep{zhu2025internvl3} & 37.5 & 28.9 & 36.9 & 45.6 \\
            InternVL3-8B~\citep{zhu2025internvl3} & 41.5 & 36.5 & 38.1 & 53.6 \\
            Qwen2.5-VL-3B~\citep{bai2025qwen2} & 37.6 & 33.5 & 35.9 & 44.8 \\
            Qwen2.5-VL-7B~\citep{bai2025qwen2} & 36.0 & \cellcolor{secondcolor}{37.0} & 32.3 & 44.0 \\
            Qwen3-VL-2B~\citep{Qwen3-VL} & 34.5 & 32.5 & 31.6 & 42.8 \\
            Qwen3-VL-8B~\citep{Qwen3-VL} & 29.4 & 29.5 & 28.6 & 31.2 \\
            
            \midrule
            \rowcolor{line-gray}\multicolumn{5}{l}{\textbf{Open-source Spatial Intelligence Models}} \\
            MindCube-3B~\citep{yin2025spatial} & \cellcolor{secondcolor}{51.7} & 34.0 & \cellcolor{secondcolor}{51.0} & \cellcolor{secondcolor}{67.6} \\
            SpatialLadder-3B~\citep{li2025spatialladder} & 43.4 & 35.0 & 43.2 & 50.8 \\
            SpaceR-7B~\citep{ouyang2025spacer} & 37.9 & 35.0 & 34.2 & 49.2 \\
            ViLaSR-7B~\citep{wu2025reinforcing} & 35.0 & 35.5 & 31.0 & 44.4 \\
            VST-3B-SFT~\citep{yang2025visual} & 35.9 & 32.0 & 34.9 & 41.6 \\
            VST-7B-SFT~\citep{yang2025visual} & 39.7 & \cellcolor{secondcolor}{37.0} & 35.9 & 50.8 \\
            Cambrian-S-3B~\citep{yang2025cambriansspatialsupersensingvideo} & 32.5 & 27.0 & 33.2 & 35.2 \\
            Spatial-MLLM-4B~\citep{wu2025spatialmllmboostingmllmcapabilities} & 33.4 & 39.0 & 30.5 & 36.0 \\
            
            \midrule
            \rowcolor{line-gray}\multicolumn{5}{l}{\textbf{Ours}} \\
            \textbf{Spatial-TTT-2B}  & \cellcolor{bestcolor}{\textbf{76.2}}& \cellcolor{bestcolor}{\textbf{55.5}}& \cellcolor{bestcolor}{\textbf{74.0}}& \cellcolor{bestcolor}{\textbf{89.8}} \\
            
            \bottomrule
        \end{tabular}%
    }
    \vspace{-0.5cm}
\end{wraptable}

\noindent \textbf{Results on MindCube Benchmark.}
We further evaluate our model and the baseline models on MindCube~\citep{yin2025spatial}, a benchmark that pairs multi-view image groups with spatial reasoning questions to assess fine-grained spatial capabilities. MindCube specifically tests (1) cross-view object consistency and (2) reasoning about occluded or unseen elements under changing camera viewpoints. Following common practice~\citep{cai2025scalingspatialintelligencemultimodal}, we use MindCube-Tiny as evaluation set, which is a diagnostic subset sampled from MindCube. Mindcube-tiny contains 1{,}050 questions in total (600 \textsc{Among}, 250 \textsc{Around}, and 200 \textsc{Rotation}). We report accuracy (ACC) for all question types. 

The results are summarized in Table~\ref{tab:mindcube}.
As shown, Spatial-TTT achieves 76.2 ACC score on MindCube-Tiny, outperforming all baseline models. By question type, it attains the best performance in \textsc{across} and \textsc{among} subsets. Relative to the strongest proprietary baseline (Gemini-3-pro; 63.9\% average) and the strongest open-source spatial model (MindCube-3B; 51.7\% average), Spatial-TTT improves by 12.3 and 24.5 percentage points, respectively. These results indicate that Spatial-TTT provides stronger spatial reasoning under viewpoint changes and occlusions.

\subsection{Streaming Spatial Sensing Results} 
To assess models’ continual spatial sensing under long-horizon, streaming-style video, we benchmark our model and baselines on VSI-SUPER-Recall (VSR) and VSI-SUPER-Count (VSC)~\citep{yang2025cambriansspatialsupersensingvideo}. Both benchmarks target spatial supersensing in arbitrarily long egocentric videos by requiring models to continuously accumulate evidence as new objects appear under changing viewpoints and to maintain long-term spatiotemporal memory for subsequent queries. Specifically, VSC requires models to count objects across extended sequences, whereas VSR uses a multiple-choice setting to test whether models can recall the temporal order of inserted objects over even longer videos. Following the benchmark protocol, we report mean recall accuracy (MRA) for VSC and accuracy (ACC) for VSR. As shown in Table~\ref{tab:vsi-super}, Spatial-TTT achieves competitive performance on VSI-SUPER-Recall across different video lengths, while significantly outperforming all baselines on VSI-SUPER-Count. 
Long-video understanding models (\eg, MovieChat~\citep{song2024moviechat} and Flash-VStream~\citep{zhang2024flash}) consistently underperform across all video durations, suggesting that strong long-context modeling alone is insufficient for continual spatial sensing without robust spatial reasoning capabilities. General purpose and spatial intelligence models (\eg, Qwen3-VL~\citep{Qwen3-VL} and Cambrian-S~\citep{yang2025cambriansspatialsupersensingvideo}) achieve relatively competitive results on shorter videos, but their performance collapses on longer sequences. This degradation is largely attributable to practical limitations in processing extended inputs, such as exceeding context budgets or running into out-of-memory, which prevents them from continuously integrating new observations and retaining long-term spatiotemporal evidence. In contrast, Spatial-TTT maintains stable performance as video length increases by performing online updates that incrementally integrate new observations, enabling continual accumulation and retention of long-term spatiotemporal evidence.
\begin{table*}[t]
\centering
\caption{\textbf{Results on VSI-SUPER-Recall (VSR) and VSI-SUPER-Count (VSC)~\citep{yang2025cambriansspatialsupersensingvideo}.} We report the average score on the 10-120 minute subset of VSR and the 10–120 minute subset of VSC. For all subsets, videos are sampled at 1 fps. Qwen3-VL-2B and Cambrian-S-7B run out of memory (OOM) on an 80GB-GPU for videos longer than 120 minutes; therefore, their scores are reported as 0.}
\label{tab:vsi-super}
\setlength{\tabcolsep}{4pt}
\renewcommand{\arraystretch}{1.2}
\resizebox{1.0\linewidth}{!}{
\begin{tabular}{l*{9}{c}}
\toprule
\multirow{2}{*}{\textbf{Models}} &
\multicolumn{4}{c}{\textbf{VSI-SUPER-Recall}} &
\multicolumn{4}{c}{\textbf{VSI-SUPER-Count}} \\
\cmidrule(lr){2-5}\cmidrule(lr){6-10}
&
\makecell{\textbf{10min}} &
\makecell{\textbf{30min}} &
\makecell{\textbf{60min}} &
\makecell{\textbf{120min}} &
\makecell{\textbf{10min}} &
\makecell{\textbf{30min}} &
\makecell{\textbf{60min}} &
\makecell{\textbf{120min}}\\
\midrule
\rowcolor{line-gray}\multicolumn{10}{l}{\textbf{General Models and Spatial Intelligence Models}} \\

Qwen3-VL-2B~\citep{Qwen3-VL}   & 35.0 & 30.0  & \cellcolor{bestcolor}\textbf{28.3} & 0.0  & \cellcolor{secondcolor}{0.8} & 0.0 & 0.0 & 0.0\\
Cambrian-S-7B~\citep{yang2025cambriansspatialsupersensingvideo}  & \cellcolor{bestcolor}\textbf{38.3} & \cellcolor{bestcolor}\textbf{35.0} & 6.0 & 0.0 & 0.6 & 0.0 & 0.0 & 0.0\\

\midrule
\rowcolor{line-gray}\multicolumn{10}{l}{\textbf{Long-video Understanding Models}} \\
MovieChat~\citep{song2024moviechat} & 18.3 & 21.7 & 16.7 & 26.7 & 0.0 & 0.0 & 0.0 & 0.0\\
Flash-VStream~\citep{zhang2024flash}  & \cellcolor{secondcolor}28.3 & 33.3 & 23.3 & 28.3 & 0.0 & 0.0 & 0.0 & 0.0\\

\midrule
\rowcolor{line-gray}\multicolumn{10}{l}{\textbf{Ours}} \\
\textbf{Spatial-TTT-2B}  & \cellcolor{bestcolor}\textbf{38.3} & \cellcolor{bestcolor}\textbf{35.0} & \cellcolor{bestcolor}\textbf{28.3} & \cellcolor{bestcolor}\textbf{30.0} & \cellcolor{bestcolor}\textbf{31.8} & \cellcolor{bestcolor}\textbf{45.6} & \cellcolor{bestcolor}\textbf{36.2} & \cellcolor{bestcolor}\textbf{38.4} \\

\bottomrule
\end{tabular}
}
\end{table*}

\subsection{Ablation Study and Analysis}
In Table~\ref{tab:vsibench_ablation}, we report ablation results of Spatial-TTT on VSI-Bench~\citep{Yang_2025_CVPR}. As shown, the full model achieves the best overall performance (64.4 Avg.), outperforming all ablated variants on both the Numerical and Multiple-Choice subsets, validating the effectiveness of our proposed components. Following, we provide a component-wise analysis.

\noindent \textbf{Spatial-predictive mechanism.}
Replacing the depth-wise 3D spatiotemporal convolution with identity projections (\textit{w/o SP-Mechanism}) drops Avg. from 64.4 to 62.1, with a larger decline on Numerical questions (64.0 to 60.7). This supports that injecting local spatiotemporal inductive bias stabilizes fast-weight updates and benefits metric-level spatial reasoning.

\noindent \textbf{Dense scene-description supervision.}
As shown, removing dense scene description data (\textit{w/o Dense Data}) reduces the overall Avg. from 64.4 to 61.3. On the Numerical and Multiple-Choice subsets, the scores drop by 3.0 and 3.3 points, respectively. These results indicate that dense, scene-level supervision provides stronger training signals for learning effective online update dynamics and for retaining globally useful 3D evidence over long video streams.

\noindent \textbf{Hybrid TTT architecture.}
Eliminating self-attention anchor layers (\textit{w/o Hybrid Arch}) causes the largest degradation on average score (from 64.4 to 53.9), especially on Multiple-Choice (from 64.8 to 52.4). These results highlights the necessity of hybrid interleaving to retain cross-modal alignment and global-context reasoning while scaling to long-horizon videos.
\begin{wraptable}{r}{7.6cm} 
    \centering
    \vspace{-0.2cm} 
    \renewcommand{\arraystretch}{1.2} 
    \caption{\textbf{Ablations of Spatial-TTT on VSI-Bench~\citep{Yang_2025_CVPR}.}
    ``w/o SP-Mechanism '' denotes replacing spatial-predictive conv with identity projections.
    ``w/o Dense Data'' denotes training without dense scene-description data.
    ``w/o Hybrid Arch'' denotes pure TTT architecture.}
    \label{tab:vsibench_ablation}
    \resizebox{\linewidth}{!}{%
        \begin{tabular}{lccc}
            \toprule
            \textbf{Setting} & \textbf{Numerical} & \textbf{Multiple-Choice} & \textbf{Avg.} \\
            \midrule
            Spatial-TTT & \cellcolor{bestcolor}{\textbf{64.0}} &\cellcolor{bestcolor}{\textbf{64.8}} & \cellcolor{bestcolor}{\textbf{64.4}} \\
            w/o SP-Mechanism               &  60.7 & \cellcolor{secondcolor}{63.4} & \cellcolor{secondcolor}{62.1} \\
            w/o Dense Data         & \cellcolor{secondcolor}{61.0} & 61.5 & 61.3 \\
            w/o Hybrid Arch             & 55.4 & 52.4 & 53.9 \\
            \bottomrule
        \end{tabular}%
    }
    \vspace{-1cm}
\end{wraptable}

\noindent \textbf{Analysis for memory usage and theoretical TFLOPs.}
We further analyze decoding memory usage and theoretical TFLOPs across varying input lengths. As shown in Table~\ref{tab:mem_flops}, Spatial-MLLM~\citep{wu2025spatialmllmboostingmllmcapabilities}, which introduces an explicit geometry encoder, exhibits prohibitive scaling—its computational cost grows super-linearly with frame count, and it runs out of memory (OOM) beyond 256 frames, making it impractical for streaming video scenarios.
For the general-purpose baseline Qwen3-VL-2B~\citep{Qwen3-VL}, while starting with comparable resource usage at short contexts, both its memory and TFLOPs scale with the quadratic complexity inherent to standard Transformer attention. 
In contrast, our Spatial-TTT benefits from linear-complexity attention, where doubling the input length results in approximately doubled computation—closely tracking theoretical linear scaling. At 1024 frames, Spatial-TTT achieves over 40\% reduction in both TFLOPs and memory compared to Qwen3-VL-2B. Crucially, this efficiency gap widens as context length grows, making our approach particularly well-suited for streaming spatial intelligence.

\begin{table*}[t]
\centering
\caption{\textbf{Peak memory usage (GB) and TFLOPs per forward pass for varying input lengths.} The results are measured at 352×480. For fair comparison, we select representative models with comparable base model sizes: Qwen3-VL-2B~\citep{Qwen3-VL} as the general-purpose MLLM baseline, and Spatial-MLLM-4B~\citep{wu2025spatialmllmboostingmllmcapabilities} as the geometry-augmented spatial VLM baseline. Spatial-MLLM-4B ran out of GPU memory at 512 and 1024 frames on our test hardware (reported as OOM), TFLOPs are therefore not reported.}
\label{tab:mem_flops}

\setlength{\tabcolsep}{4pt}
\renewcommand{\arraystretch}{1.2}

\resizebox{1.0\linewidth}{!}{
\begin{tabular}{lcccccccc}
\toprule

\multirow{2}{*}{\textbf{Models}} &
\multicolumn{2}{c}{\textbf{128 frames}} &
\multicolumn{2}{c}{\textbf{256 frames}} &
\multicolumn{2}{c}{\textbf{512 frames}} &
\multicolumn{2}{c}{\textbf{1024 frames}} \\

\cmidrule(lr){2-3}
\cmidrule(lr){4-5}
\cmidrule(lr){6-7}
\cmidrule(lr){8-9}

& MEM & TFLOPs
& MEM & TFLOPs
& MEM & TFLOPs
& MEM & TFLOPs \\

\midrule

Qwen3-VL-2B~\citep{Qwen3-VL}
& 6.2 & 75.9
& 8.3 & 179.9
& 12.6 & 473.9
& 21.2 & 1403.1 \\

Spatial-MLLM-4B~\citep{wu2025spatialmllmboostingmllmcapabilities}
& 25.9 & 1698.8
& 41.8 & 6002.1
& OOM & N/A
& OOM & N/A \\

\textbf{Spatial-TTT-2B (Ours)}
& 6.2 & 74.3
& 7.0 & 156.2
& 8.4 & 341.9
& 11.9 & 799.4 \\

\bottomrule
\end{tabular}
}
\vspace{-0.7cm}
\end{table*}

\section{Conclusion}
In this paper, we present Spatial-TTT, a novel framework for streaming visual-based spatial intelligence that leverages test-time training to maintain adaptive fast weights as compact memory for accumulating 3D evidence from long-horizon video streams. We employ a hybrid architecture that interleaves TTT layers with self-attention anchor layers to preserve pretrained knowledge while enabling efficient long-context processing, further enhanced by large-chunk updates and sliding-window attention for practical scalability. Beyond efficient architecture design, we further promote spatial awareness by introducing a spatial-predictive mechanism with 3D spatiotemporal convolutions to inject local neighborhood inductive bias, and construct a dense scene-description dataset to provide rich supervision for learning effective fast-weight update dynamics. Extensive experiments demonstrate that Spatial-TTT achieves state-of-the-art performance across spatial benchmarks. We hope that Spatial-TTT provides a promising direction for building MLLMs with persistent spatial memory, enabling more robust and scalable spatial intelligence in real-world applications.
\newpage

\renewcommand{\refname}{References}
\renewcommand{\bibname}{References}
\renewcommand{\bibsection}{\section*{\raggedright \Large References}}
\makeatother

\bibliographystyle{abbrvnat}
\bibliography{egbib}

@InProceedings{Yang_2025_CVPR,
  author    = {Jihan Yang and Shusheng Yang and Anjali W. Gupta and Rilyn Han and Li Fei-Fei and Saining Xie},
  title     = {Thinking in Space: How Multimodal Large Language Models See, Remember, and Recall Spaces},
  booktitle = {Proceedings of the IEEE/CVF Conference on Computer Vision and Pattern Recognition (CVPR)},
  year      = {2025},
  month     = {June},
  pages     = {10632-10643}
}

@article{huang2024rekep,
  title={Rekep: Spatio-temporal reasoning of relational keypoint constraints for robotic manipulation},
  author={Huang, Wenlong and Wang, Chen and Li, Yunzhu and Zhang, Ruohan and Fei-Fei, Li},
  journal={arXiv preprint arXiv:2409.01652},
  year={2024}
}

@article{wei2025spatial,
  title={Spatial-aware Vision Language Model for Autonomous Driving},
  author={Wei, Weijie and Luo, Zhipeng and Feng, Ling and Liong, Venice Erin},
  journal={arXiv preprint arXiv:2512.24331},
  year={2025}
}

@inproceedings{li2023blip,
  title={Blip-2: Bootstrapping language-image pre-training with frozen image encoders and large language models},
  author={Li, Junnan and Li, Dongxu and Savarese, Silvio and Hoi, Steven},
  booktitle={International conference on machine learning},
  pages={19730--19742},
  year={2023},
  organization={PMLR}
}

@article{feng2025vica,
  title={Towards visuospatial cognition via hierarchical fusion of visual experts},
  author={Feng, Qi},
  journal={arXiv preprint arXiv:2505.12363},
  year={2025}
}

@article{deng2025internspatial,
  title={Internspatial: A comprehensive dataset for spatial reasoning in vision-language models},
  author={Deng, Nianchen and Gu, Lixin and Ye, Shenglong and He, Yinan and Chen, Zhe and Li, Songze and Wang, Haomin and Wei, Xingguang and Yang, Tianshuo and Dou, Min and others},
  journal={arXiv preprint arXiv:2506.18385},
  year={2025}
}

@article{bai2025qwen2,
  title={Qwen2. 5-vl technical report},
  author={Bai, Shuai and Chen, Keqin and Liu, Xuejing and Wang, Jialin and Ge, Wenbin and Song, Sibo and Dang, Kai and Wang, Peng and Wang, Shijie and Tang, Jun and others},
  journal={arXiv preprint arXiv:2502.13923},
  year={2025}
}

@article{hurst2024gpt,
  title={Gpt-4o system card},
  author={Hurst, Aaron and Lerer, Adam and Goucher, Adam P and Perelman, Adam and Ramesh, Aditya and Clark, Aidan and Ostrow, AJ and Welihinda, Akila and Hayes, Alan and Radford, Alec and others},
  journal={arXiv preprint arXiv:2410.21276},
  year={2024}
}

@article{guo2025seed1,
  title={Seed1. 5-vl technical report},
  author={Guo, Dong and Wu, Faming and Zhu, Feida and Leng, Fuxing and Shi, Guang and Chen, Haobin and Fan, Haoqi and Wang, Jian and Jiang, Jianyu and Wang, Jiawei and others},
  journal={arXiv preprint arXiv:2505.07062},
  year={2025}
}

@article{team2024gemini,
  title={Gemini 1.5: Unlocking multimodal understanding across millions of tokens of context},
  author={Team, Gemini and Georgiev, Petko and Lei, Ving Ian and Burnell, Ryan and Bai, Libin and Gulati, Anmol and Tanzer, Garrett and Vincent, Damien and Pan, Zhufeng and Wang, Shibo and others},
  journal={arXiv preprint arXiv:2403.05530},
  year={2024}
}

@article{yang2025visual,
  title={Visual spatial tuning},
  author={Yang, Rui and Zhu, Ziyu and Li, Yanwei and Huang, Jingjia and Yan, Shen and Zhou, Siyuan and Liu, Zhe and Li, Xiangtai and Li, Shuangye and Wang, Wenqian and others},
  journal={arXiv preprint arXiv:2511.05491},
  year={2025}
}

@inproceedings{yin2025spatial,
  title={Spatial mental modeling from limited views},
  author={Yin, Baiqiao and Wang, Qineng and Zhang, Pingyue and Zhang, Jianshu and Wang, Kangrui and Wang, Zihan and Zhang, Jieyu and Chandrasegaran, Keshigeyan and Liu, Han and Krishna, Ranjay and others},
  booktitle={Structural Priors for Vision Workshop at ICCV'25},
  year={2025}
}

@article{zhu2025internvl3,
  title={Internvl3: Exploring advanced training and test-time recipes for open-source multimodal models},
  author={Zhu, Jinguo and Wang, Weiyun and Chen, Zhe and Liu, Zhaoyang and Ye, Shenglong and Gu, Lixin and Tian, Hao and Duan, Yuchen and Su, Weijie and Shao, Jie and others},
  journal={arXiv preprint arXiv:2504.10479},
  year={2025}
}

@InProceedings{Li_2025_ICCV,
    author    = {Li, Yun and Zhang, Yiming and Lin, Tao and Liu, XiangRui and Cai, Wenxiao and Liu, Zheng and Zhao, Bo},
    title     = {STI-Bench: Are MLLMs Ready for Precise Spatial-Temporal World Understanding?},
    booktitle = {Proceedings of the IEEE/CVF International Conference on Computer Vision (ICCV)},
    month     = {October},
    year      = {2025},
    pages     = {25332-25342}
}

@misc{yang2025cambriansspatialsupersensingvideo,
      title={Cambrian-S: Towards Spatial Supersensing in Video}, 
      author={Shusheng Yang and Jihan Yang and Pinzhi Huang and Ellis Brown and Zihao Yang and Yue Yu and Shengbang Tong and Zihan Zheng and Yifan Xu and Muhan Wang and Daohan Lu and Rob Fergus and Yann LeCun and Li Fei-Fei and Saining Xie},
      year={2025},
      eprint={2511.04670},
      archivePrefix={arXiv},
      primaryClass={cs.CV},
      url={https://arxiv.org/abs/2511.04670}, 
}

@inproceedings{sun2024learning,
  title={Learning to (Learn at Test Time): {RNN}s with Expressive Hidden States},
  author={Sun, Yu and Li, Xinhao and Dalal, Karan and Xu, Jiarui and Vikram, Arjun and Zhang, Genghan and Dubois, Yann and Chen, Xinlei and Wang, Xiaolong and Koyejo, Sanmi and Hashimoto, Tatsunori and Guestrin, Carlos},
  booktitle={Proceedings of the 41st International Conference on Machine Learning (ICML)},
  year={2024},
  url={https://arxiv.org/abs/2407.04620}
}

@InProceedings{Daxberger_2025_ICCV,
  author    = {Daxberger, Erik and Wenzel, Nina and Griffiths, David and Gang, Haiming and Lazarow, Justin and Kohavi, Gefen and Kang, Kai and Eichner, Marcin and Yang, Yinfei and Dehghan, Afshin and Grasch, Peter},
  title     = {MM-Spatial: Exploring 3D Spatial Understanding in Multimodal LLMs},
  booktitle = {Proceedings of the IEEE/CVF International Conference on Computer Vision (ICCV)},
  month     = {October},
  year      = {2025},
  pages     = {7395--7408}
}

@article{wu2025spatialmllmboostingmllmcapabilities,
    title={Spatial-MLLM: Boosting MLLM Capabilities in Visual-based Spatial Intelligence},
    author={Wu, Diankun  and Liu, Fangfu and Hung, Yi-Hsin and Duan, Yueqi},
    journal={arXiv preprint arXiv:2505.23747},
    year={2025}
}

@article{zhang2025ttt,
  title={Test-Time Training Done Right},
  author={Zhang, Tianyuan and Bi, Sai and Hong, Yicong and Zhang, Kai and Luan, Fujun and Yang, Songlin and Sunkavalli, Kalyan and Freeman, William T. and Tan, Hao},
  journal={arXiv preprint arXiv:2505.23884},
  year={2025},
  url={https://arxiv.org/abs/2505.23884}
}

@article{sun2023retentive,
  title={Retentive network: A successor to transformer for large language models},
  author={Sun, Yutao and Dong, Li and Huang, Shaohan and Ma, Shuming and Xia, Yuqing and Xue, Jilong and Wang, Jianyong and Wei, Furu},
  journal={arXiv preprint arXiv:2307.08621},
  year={2023}
}

@article{tandon2025endtoend,
  title={End-to-End Test-Time Training for Long Context},
  author={Tandon, Arnuv and Dalal, Karan and Li, Xinhao and Koceja, Daniel and R{\o}d, Marcel and Buchanan, Sam and Wang, Xiaolong and Leskovec, Jure and Koyejo, Sanmi and Hashimoto, Tatsunori and Guestrin, Carlos and McCaleb, Jed and Choi, Yejin and Sun, Yu},
  journal={arXiv preprint arXiv:2512.23675},
  year={2025},
  url={https://arxiv.org/abs/2512.23675}
}

@article{jordan2024muon,
  title={Muon: An optimizer for hidden layers in neural networks},
  author={Jordan, Keller and Jin, Yuchen and Boza, Vlado and You, Jiacheng and Cesista, Franz and Newhouse, Laker and Bernstein, Jeremy},
  journal={Cited on},
  pages={10},
  year={2024}
}

@inproceedings{sun2020test,
  title={Test-time training with self-supervision for generalization under distribution shifts},
  author={Sun, Yu and Wang, Xiaolong and Liu, Zhuang and Miller, John and Efros, Alexei and Hardt, Moritz},
  booktitle={International Conference on Machine Learning},
  pages={9229--9248},
  year={2020},
  organization={PMLR}
}

@misc{cai2025scalingspatialintelligencemultimodal,
      title={Scaling Spatial Intelligence with Multimodal Foundation Models}, 
      author={Zhongang Cai and Ruisi Wang and Chenyang Gu and Fanyi Pu and Junxiang Xu and Yubo Wang and Wanqi Yin and Zhitao Yang and Chen Wei and Qingping Sun and Tongxi Zhou and Jiaqi Li and Hui En Pang and Oscar Qian and Yukun Wei and Zhiqian Lin and Xuanke Shi and Kewang Deng and Xiaoyang Han and Zukai Chen and Xiangyu Fan and Hanming Deng and Lewei Lu and Liang Pan and Bo Li and Ziwei Liu and Quan Wang and Dahua Lin and Lei Yang},
      year={2025},
      eprint={2511.13719},
      archivePrefix={arXiv},
      primaryClass={cs.CV},
      url={https://arxiv.org/abs/2511.13719}, 
}

@inproceedings{snell2024scaling,
  title={Scaling LLM Test-Time Compute Optimally Can be More Effective than Scaling Parameters},
  author={Snell, Charlie and Jaeckle, Luke and Li, Raymond and Kumar, Aviral and Levine, Sergey},
  booktitle={Proceedings of the 41st International Conference on Machine Learning (ICML)},
  year={2024}
}

@article{deepseek2025r1,
  title={DeepSeek-R1: Incentivizing Reasoning Capability in LLMs via Reinforcement Learning},
  author={DeepSeek-AI and Guo, Daya and Yang, Dejian and Zhang, Haowei and Song, Junxiao and Wang, Peiyi and others},
  journal={Nature},
  volume={645},
  pages={633--638},
  year={2025},
  publisher={Nature Publishing Group},
  doi={10.1038/s41586-025-09422-z}
}

@techreport{openai2024o1,
  title={OpenAI o1 System Card},
  author={{OpenAI}},
  institution={OpenAI},
  year={2024},
  url={https://openai.com/index/openai-o1-system-card/},
  note={Technical Report}
}

@article{Liang_2024,
   title={A Comprehensive Survey on Test-Time Adaptation Under Distribution Shifts},
   volume={133},
   ISSN={1573-1405},
   url={http://dx.doi.org/10.1007/s11263-024-02181-w},
   DOI={10.1007/s11263-024-02181-w},
   number={1},
   journal={International Journal of Computer Vision},
   publisher={Springer Science and Business Media LLC},
   author={Liang, Jian and He, Ran and Tan, Tieniu},
   year={2024},
   month=jul, pages={31–64} }

@inproceedings{akyurek2025surprising,
  title={The Surprising Effectiveness of Test-Time Training for Few-Shot Learning},
  author={Aky{\"u}rek, Ekin and Damani, Mehul and Zweiger, Adam and Qiu, Linlu and Guo, Han and Pari, Jyothish and Kim, Yoon and Andreas, Jacob},
  booktitle={Proceedings of the 42nd International Conference on Machine Learning},
  series={Proceedings of Machine Learning Research},
  volume={267},
  pages={942--963},
  year={2025},
  publisher={PMLR},
  url={https://proceedings.mlr.press/v267/akyurek25a.html}
}

@inproceedings{liu2024llava,
  title={Visual Instruction Tuning},
  author={Liu, Haotian and Li, Chunyuan and Wu, Qingyang and Lee, Yong Jae},
  booktitle={Proceedings of the Neural Information Processing Systems (NeurIPS)},
  year={2023}
}

@inproceedings{li2024manipllm,
  title={{ManipLLM}: Embodied Multimodal Large Language Model for Object-Centric Robotic Manipulation},
  author={Li, Xiaoqi and Zhang, Mingxu and others},
  booktitle={Proceedings of the IEEE/CVF Conference on Computer Vision and Pattern Recognition (CVPR)},
  year={2024}
}

@inproceedings{hong20233dllm,
  title={{3D-LLM}: Injecting the 3D World into Large Language Models},
  author={Hong, Yining and Zhen, Haoyu and others},
  booktitle={Proceedings of the Neural Information Processing Systems (NeurIPS)},
  year={2023}
}

@inproceedings{chen2024spatialvlm,
  title={Spatial-VLM: Endowing Vision-Language Models with Spatial Reasoning Capabilities},
  author={Chen, Boyuan and Xu, Zhuo and Kirmani, Sean and Ichter, Brian and Driess, Danny and Florence, Pete and Sadigh, Dorsa and Guibas, Leonidas and Fei-Fei, Li},
  booktitle={Proceedings of the IEEE/CVF Conference on Computer Vision and Pattern Recognition (CVPR)},
  year={2024}
}

@inproceedings{driess2023palme,
  title={PaLM-E: An Embodied Multimodal Language Model},
  author={Driess, Danny and Xia, Fei and Sajjadi, Mehdi S. M. and Lynch, Corey and Chowdhery, Aakanksha and Ichter, Brian and others},
  booktitle={International Conference on Machine Learning (ICML)},
  year={2023}
}

@inproceedings{hu2023uniad,
  title={Planning-oriented Autonomous Driving},
  author={Hu, Yihan and Yang, Jiazhi and Chen, Li and Li, Keyu and Sima, Chonghao and Zhu, Xizhou and Chai, Siqi and Du, Senyao},
  booktitle={Proceedings of the IEEE/CVF Conference on Computer Vision and Pattern Recognition (CVPR)},
  year={2023}
}

@inproceedings{grauman2022ego4d,
  title={Ego4D: Around the World in 3,000 Hours of Egocentric Video},
  author={Grauman, Kristen and Westbury, Andrew and Byrne, Eugene and Chavis, Zachary and Furnari, Antonino and Girdhar, Rohit and others},
  booktitle={Proceedings of the IEEE/CVF Conference on Computer Vision and Pattern Recognition (CVPR)},
  year={2022}
}

@inproceedings{li2020streaming,
  title={Towards Streaming Perception},
  author={Li, Mengtian and Wang, Yu-Xiong and Ramanan, Deva},
  booktitle={European Conference on Computer Vision (ECCV)},
  year={2020}
}

@article{fu2023mme,
  title={MME: A Comprehensive Evaluation Benchmark for Multimodal Large Language Models},
  author={Fu, Chaoyou and Chen, Peixian and Shen, Yunhang and Qin, Yulei and Zhang, Mengxu and Lin, Xu and others},
  journal={arXiv preprint arXiv:2306.13394},
  year={2023}
}

@inproceedings{lin2023videollava,
  title={Video-LLaVA: Learning United Visual Representation by Alignment Before Projection},
  author={Lin, Bin and Zhu, Bin and Ye, Yang and Ning, Munan and Jin, Peng and Yuan, Li},
  booktitle={arXiv preprint arXiv:2311.10122},
  year={2023}
}

@inproceedings{vaswani2017attention,
  title={Attention Is All You Need},
  author={Vaswani, Ashish and Shazeer, Noam and Parmar, Niki and Uszkoreit, Jakob and Jones, Llion and Gomez, Aidan N and Kaiser, {\L}ukasz and Polosukhin, Illia},
  booktitle={Advances in Neural Information Processing Systems (NeurIPS)},
  year={2017}
}

@inproceedings{sun2024learningtolearn,
  title={Learning to (Learn at Test Time): RNNs with Expressive Hidden States},
  author={Sun, Yu and Dong, Xupu and Menon, Shakul and Schwab, Dennis and Kolter, J. Zico},
  booktitle={arXiv preprint arXiv:2407.04620},
  year={2024}
}

@article{fan2025vlm,
  title={VLM-3R: Vision-Language Models Augmented with Instruction-Aligned 3D Reconstruction},
  author={Fan, Zhiwen and Zhang, Jian and Li, Renjie and Zhang, Junge and Chen, Runjin and Hu, Hezhen and Wang, Kevin and Qu, Huaizhi and Wang, Dilin and Yan, Zhicheng and others},
  journal={arXiv preprint arXiv:2505.20279},
  year={2025}
}

@inproceedings{jia2024sceneverse,
  title={Sceneverse: Scaling 3d vision-language learning for grounded scene understanding},
  author={Jia, Baoxiong and Chen, Yixin and Yu, Huangyue and Wang, Yan and Niu, Xuesong and Liu, Tengyu and Li, Qing and Huang, Siyuan},
  booktitle={European Conference on Computer Vision},
  pages={289--310},
  year={2024},
  organization={Springer}
}

@article{Qwen3-VL,
      title={Qwen3-VL Technical Report}, 
      author={Shuai Bai and Yuxuan Cai and Ruizhe Chen and Keqin Chen and Xionghui Chen and Zesen Cheng and Lianghao Deng and Wei Ding and Chang Gao and Chunjiang Ge and Wenbin Ge and Zhifang Guo and Qidong Huang and Jie Huang and Fei Huang and Binyuan Hui and Shutong Jiang and Zhaohai Li and Mingsheng Li and Mei Li and Kaixin Li and Zicheng Lin and Junyang Lin and Xuejing Liu and Jiawei Liu and Chenglong Liu and Yang Liu and Dayiheng Liu and Shixuan Liu and Dunjie Lu and Ruilin Luo and Chenxu Lv and Rui Men and Lingchen Meng and Xuancheng Ren and Xingzhang Ren and Sibo Song and Yuchong Sun and Jun Tang and Jianhong Tu and Jianqiang Wan and Peng Wang and Pengfei Wang and Qiuyue Wang and Yuxuan Wang and Tianbao Xie and Yiheng Xu and Haiyang Xu and Jin Xu and Zhibo Yang and Mingkun Yang and Jianxin Yang and An Yang and Bowen Yu and Fei Zhang and Hang Zhang and Xi Zhang and Bo Zheng and Humen Zhong and Jingren Zhou and Fan Zhou and Jing Zhou and Yuanzhi Zhu and Ke Zhu},
	  journal={arXiv preprint arXiv:2511.21631},
      year={2025}
}

@inproceedings{xu2023pointllm,
  title={PointLLM: Empowering Large Language Models to Understand Point Clouds},
  author={Xu, Runsen and Wang, Xiangtai and Zhang, Tai and others},
  booktitle={CVPR},
  year={2024}
}

@inproceedings{li2024mvbench,
  title={MVBench: A Comprehensive Multi-modal Video Understanding Benchmark},
  author={Li, Kunchang and Wang, Yali and He, Yinan and others},
  booktitle={CVPR},
  year={2024}
}

@inproceedings{wang2021tent,
  title={Tent: Fully Test-Time Adaptation by Entropy Minimization},
  author={Wang, Dequan and Shelhamer, Evan and Liu, Shaoteng and others},
  booktitle={ICLR},
  year={2021}
}

@inproceedings{gandelsman2022test,
  title={Test-Time Training with Masked Autoencoders},
  author={Gandelsman, Yossi and Sun, Yu and Chen, Xinlei and Efros, Alexei A},
  booktitle={NeurIPS},
  year={2022}
}

@inproceedings{ba2016using,
  title={Using Fast Weights to Attend to the Recent Past},
  author={Ba, Jimmy and Hinton, Geoffrey E and Mnih, Volodymyr and others},
  booktitle={NeurIPS},
  year={2016}
}

@inproceedings{shu2022tpt,
  title={Test-Time Prompt Tuning for Zero-Shot Generalization in Vision-Language Models},
  author={Shu, Manli and Nie, Weixiao and Huang, De-An and others},
  booktitle={NeurIPS},
  year={2022}
}

@inproceedings{dai2017scannet,
  title={Scannet: Richly-annotated 3d reconstructions of indoor scenes},
  author={Dai, Angela and Chang, Angel X and Savva, Manolis and Halber, Maciej and Funkhouser, Thomas and Nie{\ss}ner, Matthias},
  booktitle={Proceedings of the IEEE conference on computer vision and pattern recognition},
  pages={5828--5839},
  year={2017}
}

@inproceedings{yeshwanth2023scannet++,
  title={Scannet++: A high-fidelity dataset of 3d indoor scenes},
  author={Yeshwanth, Chandan and Liu, Yueh-Cheng and Nie{\ss}ner, Matthias and Dai, Angela},
  booktitle={Proceedings of the IEEE/CVF International Conference on Computer Vision},
  pages={12--22},
  year={2023}
}

@article{baruch2021arkitscenes,
  title={Arkitscenes: A diverse real-world dataset for 3d indoor scene understanding using mobile rgb-d data},
  author={Baruch, Gilad and Chen, Zhuoyuan and Dehghan, Afshin and Dimry, Tal and Feigin, Yuri and Fu, Peter and Gebauer, Thomas and Joffe, Brandon and Kurz, Daniel and Schwartz, Arik and others},
  journal={arXiv preprint arXiv:2111.08897},
  year={2021}
}

@article{li2025spatialladder,
  title={Spatialladder: Progressive training for spatial reasoning in vision-language models},
  author={Li, Hongxing and Li, Dingming and Wang, Zixuan and Yan, Yuchen and Wu, Hang and Zhang, Wenqi and Shen, Yongliang and Lu, Weiming and Xiao, Jun and Zhuang, Yueting},
  journal={arXiv preprint arXiv:2510.08531},
  year={2025}
}

@article{wu2025reinforcing,
  title={Reinforcing spatial reasoning in vision-language models with interwoven thinking and visual drawing},
  author={Wu, Junfei and Guan, Jian and Feng, Kaituo and Liu, Qiang and Wu, Shu and Wang, Liang and Wu, Wei and Tan, Tieniu},
  journal={arXiv preprint arXiv:2506.09965},
  year={2025}
}

@article{ouyang2025spacer,
  title={SpaceR: Reinforcing MLLMs in Video Spatial Reasoning},
  author={Ouyang, Kun and Liu, Yuanxin and Wu, Haoning and Liu, Yi and Zhou, Hao and Zhou, Jie and Meng, Fandong and Sun, Xu},
  journal={arXiv preprint arXiv:2504.01805},
  year={2025}
}

@article{li2024llava,
  title={Llava-onevision: Easy visual task transfer},
  author={Li, Bo and Zhang, Yuanhan and Guo, Dong and Zhang, Renrui and Li, Feng and Zhang, Hao and Zhang, Kaichen and Zhang, Peiyuan and Li, Yanwei and Liu, Ziwei and others},
  journal={arXiv preprint arXiv:2408.03326},
  year={2024}
}

@article{zhang2024long,
  title={Long context transfer from language to vision},
  author={Zhang, Peiyuan and Zhang, Kaichen and Li, Bo and Zeng, Guangtao and Yang, Jingkang and Zhang, Yuanhan and Wang, Ziyue and Tan, Haoran and Li, Chunyuan and Liu, Ziwei},
  journal={arXiv preprint arXiv:2406.16852},
  year={2024}
}

@inproceedings{song2024moviechat,
  title={Moviechat: From dense token to sparse memory for long video understanding},
  author={Song, Enxin and Chai, Wenhao and Wang, Guanhong and Zhang, Yucheng and Zhou, Haoyang and Wu, Feiyang and Chi, Haozhe and Guo, Xun and Ye, Tian and Zhang, Yanting and others},
  booktitle={Proceedings of the IEEE/CVF Conference on Computer Vision and Pattern Recognition},
  pages={18221--18232},
  year={2024}
}

@article{zhang2024flash,
  title={Flash-vstream: Memory-based real-time understanding for long video streams},
  author={Zhang, Haoji and Wang, Yiqin and Tang, Yansong and Liu, Yong and Feng, Jiashi and Dai, Jifeng and Jin, Xiaojie},
  journal={arXiv preprint arXiv:2406.08085},
  year={2024}
}

@article{chen2025think,
  title={Think with 3d: Geometric imagination grounded spatial reasoning from limited views},
  author={Chen, Zhangquan and Zhang, Manyuan and Yu, Xinlei and Luo, Xufang and Sun, Mingze and Pan, Zihao and Feng, Yan and Pei, Peng and Cai, Xunliang and Huang, Ruqi},
  journal={arXiv preprint arXiv:2510.18632},
  year={2025}
}

@online{grok4_xai_2025,
  title   = {Grok 4},
  author  = {{xAI}},
  year    = {2025},
  month   = {7},
  day     = {9},
  url     = {https://x.ai/news/grok-4},
  urldate = {2025-09-24},
  note    = {Model announcement}
}

@article{comanici2025gemini,
  title={Gemini 2.5: Pushing the frontier with advanced reasoning, multimodality, long context, and next generation agentic capabilities},
  author={Comanici, Gheorghe and Bieber, Eric and Schaekermann, Mike and Pasupat, Ice and Sachdeva, Noveen and Dhillon, Inderjit and Blistein, Marcel and Ram, Ori and Zhang, Dan and Rosen, Evan and others},
  journal={arXiv preprint arXiv:2507.06261},
  year={2025}
}

@misc{openai_gpt5_systemcard,
  author       = {{OpenAI}},
  title        = {{GPT-5 System Card}},
  howpublished = {Technical report, OpenAI},
  note         = {Accessed: 2025-08-10},
  year         = {2025},
  month        = aug,
  day          = 7,
}

@InProceedings{schlag2021linear,
  title = 	 {Linear Transformers Are Secretly Fast Weight Programmers},
  author =       {Schlag, Imanol and Irie, Kazuki and Schmidhuber, J\"urgen},
  booktitle = 	 {Proceedings of the 38th International Conference on Machine Learning},
  pages = 	 {9355--9366},
  year = 	 {2021},
  editor = 	 {Meila, Marina and Zhang, Tong},
  volume = 	 {139},
  series = 	 {Proceedings of Machine Learning Research},
  month = 	 {18--24 Jul},
  publisher =    {PMLR},
  url = 	 {https://proceedings.mlr.press/v139/schlag21a.html}
}

@article{wang2025test,
  title={Test-time regression: a unifying framework for designing sequence models with associative memory},
  author={Wang, Ke Alexander and Shi, Jiaxin and Fox, Emily B},
  journal={arXiv preprint arXiv:2501.12352},
  year={2025}
}

@article{behrouz2024titans,
  title={Titans: Learning to memorize at test time},
  author={Behrouz, Ali and Zhong, Peilin and Mirrokni, Vahab},
  journal={arXiv preprint arXiv:2501.00663},
  year={2024}
}

@article{karami2025lattice,
  title={Lattice: Learning to efficiently compress the memory},
  author={Karami, Mahdi and Mirrokni, Vahab},
  journal={arXiv preprint arXiv:2504.05646},
  year={2025}
}

@techreport{seed20,
  title={Seed2.0 Model Card: Towards Intelligence Frontier for Real-World Complexity},
  author={{Bytedance Seed}},
  institution={Bytedance},
  year={2025},
  url={https://lf3-static.bytednsdoc.com/obj/eden-cn/lapzild-tss/ljhwZthlaukjlkulzlp/seed2/0214/Seed2.0%20Model%20Card.pdf/},
  note={Technical Report}
}

@online{gemini3,
  title   = {A new era of intelligence with Gemini 3},
  author  = {{Google}},
  url     = {https://blog.google/products-and-platforms/products/gemini/gemini-3},
}

@misc{kimiteam2026kimik25visualagentic,
      title={Kimi K2.5: Visual Agentic Intelligence}, 
      author={Kimi Team and Tongtong Bai and Yifan Bai and Yiping Bao and S. H. Cai and Yuan Cao and Y. Charles and H. S. Che and Cheng Chen and Guanduo Chen and Huarong Chen and Jia Chen and Jiahao Chen and Jianlong Chen and Jun Chen and Kefan Chen and Liang Chen and Ruijue Chen and Xinhao Chen and Yanru Chen and Yanxu Chen and Yicun Chen and Yimin Chen and Yingjiang Chen and Yuankun Chen and Yujie Chen and Yutian Chen and Zhirong Chen and Ziwei Chen and Dazhi Cheng and Minghan Chu and Jialei Cui and Jiaqi Deng and Muxi Diao and Hao Ding and Mengfan Dong and Mengnan Dong and Yuxin Dong and Yuhao Dong and Angang Du and Chenzhuang Du and Dikang Du and Lingxiao Du and Yulun Du and Yu Fan and Shengjun Fang and Qiulin Feng and Yichen Feng and Garimugai Fu and Kelin Fu and Hongcheng Gao and Tong Gao and Yuyao Ge and Shangyi Geng and Chengyang Gong and Xiaochen Gong and Zhuoma Gongque and Qizheng Gu and Xinran Gu and Yicheng Gu and Longyu Guan and Yuanying Guo and Xiaoru Hao and Weiran He and Wenyang He and Yunjia He and Chao Hong and Hao Hu and Jiaxi Hu and Yangyang Hu and Zhenxing Hu and Ke Huang and Ruiyuan Huang and Weixiao Huang and Zhiqi Huang and Tao Jiang and Zhejun Jiang and Xinyi Jin and Yu Jing and Guokun Lai and Aidi Li and C. Li and Cheng Li and Fang Li and Guanghe Li and Guanyu Li and Haitao Li and Haoyang Li and Jia Li and Jingwei Li and Junxiong Li and Lincan Li and Mo Li and Weihong Li and Wentao Li and Xinhang Li and Xinhao Li and Yang Li and Yanhao Li and Yiwei Li and Yuxiao Li and Zhaowei Li and Zheming Li and Weilong Liao and Jiawei Lin and Xiaohan Lin and Zhishan Lin and Zichao Lin and Cheng Liu and Chenyu Liu and Hongzhang Liu and Liang Liu and Shaowei Liu and Shudong Liu and Shuran Liu and Tianwei Liu and Tianyu Liu and Weizhou Liu and Xiangyan Liu and Yangyang Liu and Yanming Liu and Yibo Liu and Yuanxin Liu and Yue Liu and Zhengying Liu and Zhongnuo Liu and Enzhe Lu and Haoyu Lu and Zhiyuan Lu and Junyu Luo and Tongxu Luo and Yashuo Luo and Long Ma and Yingwei Ma and Shaoguang Mao and Yuan Mei and Xin Men and Fanqing Meng and Zhiyong Meng and Yibo Miao and Minqing Ni and Kun Ouyang and Siyuan Pan and Bo Pang and Yuchao Qian and Ruoyu Qin and Zeyu Qin and Jiezhong Qiu and Bowen Qu and Zeyu Shang and Youbo Shao and Tianxiao Shen and Zhennan Shen and Juanfeng Shi and Lidong Shi and Shengyuan Shi and Feifan Song and Pengwei Song and Tianhui Song and Xiaoxi Song and Hongjin Su and Jianlin Su and Zhaochen Su and Lin Sui and Jinsong Sun and Junyao Sun and Tongyu Sun and Flood Sung and Yunpeng Tai and Chuning Tang and Heyi Tang and Xiaojuan Tang and Zhengyang Tang and Jiawen Tao and Shiyuan Teng and Chaoran Tian and Pengfei Tian and Ao Wang and Bowen Wang and Chensi Wang and Chuang Wang and Congcong Wang and Dingkun Wang and Dinglu Wang and Dongliang Wang and Feng Wang and Hailong Wang and Haiming Wang and Hengzhi Wang and Huaqing Wang and Hui Wang and Jiahao Wang and Jinhong Wang and Jiuzheng Wang and Kaixin Wang and Linian Wang and Qibin Wang and Shengjie Wang and Shuyi Wang and Si Wang and Wei Wang and Xiaochen Wang and Xinyuan Wang and Yao Wang and Yejie Wang and Yipu Wang and Yiqin Wang and Yucheng Wang and Yuzhi Wang and Zhaoji Wang and Zhaowei Wang and Zhengtao Wang and Zhexu Wang and Zihan Wang and Zizhe Wang and Chu Wei and Ming Wei and Chuan Wen and Zichen Wen and Chengjie Wu and Haoning Wu and Junyan Wu and Rucong Wu and Wenhao Wu and Yuefeng Wu and Yuhao Wu and Yuxin Wu and Zijian Wu and Chenjun Xiao and Jin Xie and Xiaotong Xie and Yuchong Xie and Yifei Xin and Bowei Xing and Boyu Xu and Jianfan Xu and Jing Xu and Jinjing Xu and L. H. Xu and Lin Xu and Suting Xu and Weixin Xu and Xinbo Xu and Xinran Xu and Yangchuan Xu and Yichang Xu and Yuemeng Xu and Zelai Xu and Ziyao Xu and Junjie Yan and Yuzi Yan and Guangyao Yang and Hao Yang and Junwei Yang and Kai Yang and Ningyuan Yang and Ruihan Yang and Xiaofei Yang and Xinlong Yang and Ying Yang and Yi Yang and Yi Yang and Zhen Yang and Zhilin Yang and Zonghan Yang and Haotian Yao and Dan Ye and Wenjie Ye and Zhuorui Ye and Bohong Yin and Chengzhen Yu and Longhui Yu and Tao Yu and Tianxiang Yu and Enming Yuan and Mengjie Yuan and Xiaokun Yuan and Yang Yue and Weihao Zeng and Dunyuan Zha and Haobing Zhan and Dehao Zhang and Hao Zhang and Jin Zhang and Puqi Zhang and Qiao Zhang and Rui Zhang and Xiaobin Zhang and Y. Zhang and Yadong Zhang and Yangkun Zhang and Yichi Zhang and Yizhi Zhang and Yongting Zhang and Yu Zhang and Yushun Zhang and Yutao Zhang and Yutong Zhang and Zheng Zhang and Chenguang Zhao and Feifan Zhao and Jinxiang Zhao and Shuai Zhao and Xiangyu Zhao and Yikai Zhao and Zijia Zhao and Huabin Zheng and Ruihan Zheng and Shaojie Zheng and Tengyang Zheng and Junfeng Zhong and Longguang Zhong and Weiming Zhong and M. Zhou and Runjie Zhou and Xinyu Zhou and Zaida Zhou and Jinguo Zhu and Liya Zhu and Xinhao Zhu and Yuxuan Zhu and Zhen Zhu and Jingze Zhuang and Weiyu Zhuang and Ying Zou and Xinxing Zu},
      year={2026},
      eprint={2602.02276},
      archivePrefix={arXiv},
      primaryClass={cs.CL},
      url={https://arxiv.org/abs/2602.02276}, 
}

\newpage
\appendix

\section{Additional Implementation Details}

\subsection{Spatial Dataset Curation}
The first-stage dense scene-description dataset is built from object-centric 3D scene graphs provided by SceneVerse~\citep{jia2024sceneverse}, and is used to train the hybrid TTT architecture so that fast weights learn to retain comprehensive scene-level information through chunk-by-chunk updates. Each sample pairs a spatial video stream with a target description in the form of a coherent scene walkthrough, covering global context (scene type and functional setting), object categories and counts, and spatial layouts and pairwise relations. The dataset comprises approximately 16K samples in total: 3.6K from ScanNet~\citep{dai2017scannet} and 12.5K from ARKitScenes~\citep{baruch2021arkitscenes}, providing dense, high-coverage supervision that complements the sparser spatial QA signals used in the second stage.

In the second stage, we train the model on a large-scale spatial question-answering dataset, which consists of $\sim$2.5M open-sourced data and $\sim$0.5M self-collected data. For the open-sourced general spatial understanding data, we collect VSI-590K~\citep{yang2025cambriansspatialsupersensingvideo}, VLM-3R~\citep{fan2025vlm}, InternSpatial~\citep{deng2025internspatial}, ViCA~\citep{feng2025vica}. For the self-collected data, we first sample frames from the raw ScanNet~\citep{dai2017scannet, yeshwanth2023scannet++} reconstructions and assemble them into indoor-scene video sequences at 24\,fps and $640\times480$ resolution. From the scene meshes and semantic annotations we derive spatial and semantic metadata. We align each raw mesh with the provided axis-alignment matrices and convert it to a point cloud. At the room level, we estimate room extent and centroid via the alpha-shape algorithm. At the object level, we fit oriented bounding boxes (OBBs) for every valid object instance and assign semantic labels from the annotations, discarding structural elements (e.g., walls, floors) and ambiguous categories. For consistency we remap the original ScanNet semantic IDs to a consolidated 40-class indoor label set. We also compute and store the 2D projected semantic annotation for each scene video to support appearance-order reasoning. The resulting metadata per sample comprises: (i) room size and center coordinates; (ii) the 2D semantic projection of the scene video; (iii) object instances with their OBB parameters (rotation matrices, extents, and centers); and (iv) semantic labels for each object.

\subsection{Prompts Details}
\label{sec:prompts}

\subsubsection{Evaluation and Inference Prompts}
\label{sec:prompt_eval}
To ensure reproducibility, we list the detailed prompt templates used for each benchmark in Table~\ref{tab:prompts_eval}. All prompts are presented in a unified Python f-string format.

{
\small
\renewcommand{\arraystretch}{1.3}

\begin{xltabular}{\linewidth}{@{}l X@{}}
    \caption{Prompt Templates for Evaluation Benchmarks} \label{tab:prompts_eval} \\
    \toprule
    \textbf{Benchmark} & \textbf{Prompt Template (Unified Format)} \\
    \midrule
    \endfirsthead

    \multicolumn{2}{c}{{\bfseries \tablename\ \thetable{} -- continued from previous page}} \\
    \toprule
    \textbf{Benchmark} & \textbf{Prompt Template (Continued)} \\
    \midrule
    \endhead

    \midrule
    \multicolumn{2}{r}{{Continued on next page...}} \\
    \endfoot

    \bottomrule
    \endlastfoot

    \makecell[l]{VSI-Bench}
    & 1. \texttt{f"\{question\}\textbackslash nOptions:\textbackslash n\{options\}\textbackslash nAnswer with the option's letter from the given choices directly."} \newline
    2. \texttt{f"\{question\}\textbackslash nPlease answer the question using a single word or phrase."} \\
    \midrule

    MindCube
    & \texttt{f"\{input\_prompt\}"} (dataset \texttt{input\_prompt} used as-is; multiple-choice, answer with option letter.) \\
    \midrule

    \makecell[l]{VSI-SUPER-Recall}
    & \texttt{f"\{question\}\textbackslash nOptions:\textbackslash n\{options\}\textbackslash nAnswer with the option's letter from the given choices directly."} \\
    \midrule

    \makecell[l]{VSI-SUPER-Count}
    & \texttt{f"These are frames of a video.\textbackslash n\{question\}\textbackslash nPlease answer the question using a single word or phrase."} \\
\end{xltabular}
}


\subsection{Algorithms of Spatial-TTT}
\label{sec:appendix_algorithms}

For clarity, we provide the pseudocode for a single hybrid TTT layer $\ell \in \mathcal{S}$ in Algorithm~\ref{alg:spatial_ttt_prefill}. The architecture interleaves TTT layers ($\ell \in \mathcal{S}$) with standard self-attention anchor layers ($\ell \notin \mathcal{S}$) at a 3:1 ratio; anchor layers use standard causal self-attention and are omitted here for brevity.

\begin{algorithm}[!t]
\caption{Prefilling of Hybrid Layer in Spatial-TTT}
\label{alg:spatial_ttt_prefill}
\begin{algorithmic}[1]
\Require Fast weights $W$; Apply function of fast weights $f_W$; Conv3D kernel $\Theta$; chunk size $C$; SWA window $B$; learning rate $\eta$.
\State \textbf{Input:} Hidden states $X \in \mathbb{R}^{T \times d}$; video mask $\mathcal{M}_v$; grid $(t,h,w)$. \Comment{Full input sequence.}
\State $Q, K, V \leftarrow \text{QKVProj}(X)$ \Comment{Step 1: Shared QKV projections.}
\State $\tilde{Q}, \tilde{K} \leftarrow \text{QKShift}(Q, K)$ \Comment{Learnable scale/shift for TTT branch.}
\State $(\tilde{Q}_v, \tilde{K}_v, V_v) \leftarrow \mathrm{PackToVolume}(\tilde{Q}, \tilde{K}, V;\, \mathcal{M}_v, t, h, w)$ \Comment{Step 2: Spatial-predictive mechanism.}
\State $(\tilde{Q}_v, \tilde{K}_v, \tilde{V}_v) \leftarrow \mathrm{DWConv3D}_\Theta(\tilde{Q}_v, \tilde{K}_v, V_v)$ \Comment{Depth-wise spatiotemporal convolution.}
\State $(\tilde{Q}, \tilde{K}, \tilde{V}) \leftarrow \mathrm{ScatterToTokens}(\tilde{Q}_v, \tilde{K}_v, \tilde{V}_v;\, \mathcal{M}_v)$
\State Partition $Y$ into $\{Y_i\}_{i=1}^{I}$ with $Y_j\in \mathbb{R}^{C\times d}$ and tail $Y_{\mathrm{tail}}$ for $Y\in\{\tilde Q, \tilde K, \tilde V\}$ \Comment{Step 3: Chunked Muon updates.}
\For{$i = 1, \ldots, I$}
    \State $O_{i} \leftarrow f_W({Q_i})$ \Comment{Apply fast weights before update.}
    \State $G \leftarrow \mathrm{MuonUpdate}\!\left(G,\, \nabla_W \mathcal{L}(f_W(\tilde{K}_{i}),\, \tilde{V}_{i})\right)$ \Comment{Orth.\ gradient + momentum.}
    \State $W \leftarrow \mathrm{L2Norm}(W - \eta\, G)$ \Comment{Update fast weights.}
\EndFor
\State $O_{\mathrm{tail}} \leftarrow f_W(Q_\text{tail})$ \Comment{Tail: apply only.}
\State $O^\text{TTT}\leftarrow \text{Concat}(\{O_i\}_{i=1}^I,O_\text{tail})$ \Comment{Concatenate all TTT output.}
\State $O^\text{Attn} \leftarrow \mathrm{SlidingWindowAttn}(Q, K, V;\, B)$ \Comment{Step 4: SWA branch.}

\State \textbf{return} $\text{OProj}(O^\text{TTT} + O^\text{Attn})$

\end{algorithmic}
\end{algorithm}

\section{Additional Results}

\subsection{More Qualitative Comparisons}
\begin{table*}[t]
\centering
\caption{\textbf{Full ablation results on VSI-Bench~\citep{Yang_2025_CVPR}.}
``w/o SP-Mehcanism'' denotes replacing spatial-predictive 3D conv with identity projections.
``w/o Dense Data'' denotes training without dense scene data.
``w/o Hybrid'' denotes pure TTT architecture.} 
\label{tab:ablation-full}

\renewcommand{\arraystretch}{1.0} 
\setlength{\tabcolsep}{1.5pt} 

\resizebox{1.0\linewidth}{!}{
\begin{tabular}{l*{9}{c}}
\toprule
\multirow{2}{*}{\textbf{Models}} &
\multicolumn{4}{c}{\textbf{Numerical Question}} &
\multicolumn{4}{c}{\textbf{Multiple-Choice Question}} &
\multirow{2}{*}{\textbf{Avg.}} \\
\cmidrule(lr){2-5}\cmidrule(lr){6-9}
&
\makecell{\textbf{Obj. Count}} &
\makecell{\textbf{Abs. Dist}} &
\makecell{\textbf{Obj. Size}} &
\makecell{\textbf{Room Size}} &
\makecell{\textbf{Rel. Dis}} &
\makecell{\textbf{Rel. Dir}} &
\makecell{\textbf{Route Plan}} &
\makecell{\textbf{Appr. Order}} &
\multicolumn{1}{c}{} \\
\midrule
Spatial-TTT         & \cellcolor{bestcolor}\textbf{70.8} & \cellcolor{bestcolor}\textbf{47.8} & \cellcolor{bestcolor}\textbf{71.7} & \cellcolor{bestcolor}\textbf{65.9} &  \cellcolor{secondcolor}61.8 & \cellcolor{bestcolor}\textbf{73.0} &  \cellcolor{secondcolor}47.4 & \cellcolor{bestcolor}\textbf{77.0} & \cellcolor{bestcolor}\textbf{64.4} \\
w/o SP-Mechanism    & 68.7 & 43.0 & 70.3 &  \cellcolor{secondcolor}60.9 & \cellcolor{bestcolor}\textbf{62.1} &  \cellcolor{secondcolor}71.3 & 45.9 &  \cellcolor{secondcolor}74.3 &  \cellcolor{secondcolor}62.1 \\
w/o Dense Data      &  \cellcolor{secondcolor}69.7 &  \cellcolor{secondcolor}44.9 &  \cellcolor{secondcolor}70.6 & 58.8 & 59.6 & 66.7 & \cellcolor{bestcolor}\textbf{47.9} & 71.8 & 61.3 \\
w/o Hybrid          & 66.5 & 36.7 & 64.5 & 54.1 & 53.0 & 58.3 & 38.1 & 60.0 & 53.9 \\
\bottomrule
\end{tabular}
}

\end{table*}

\noindent\textbf{Full Ablation Results.} We report complete ablation results on VSI-Bench in Table~\ref{tab:ablation-full}. As shown, Spatial-TTT achieves the best overall performance (Avg.~64.4), ranking first across all numerical-question categories (Obj.~Count, Abs.~Dist, Obj.~Size, and Room~Size), and also attaining top results on key multiple-choice questions such as Rel.~Dir and Appr.~Order. 
Removing the SP-Mechanism leads to a clear performance drop, most notably on numerical questions and Route~Plan, indicating that the proposed spatial-predictive mechanism is essential for precise metric-aware spatial perception and planning. 
In contrast, removing Dense Data results in marked degradations on Room~Size, Rel.~Dir, and Appr.~Order, suggesting that dense scene supervision helps the model build stronger long-horizon spatial memory and better integrate temporally dependent observations required by sequence-dependent reasoning tasks.

\begin{figure}[!t]
    \centering
    \includegraphics[width=0.9\linewidth]{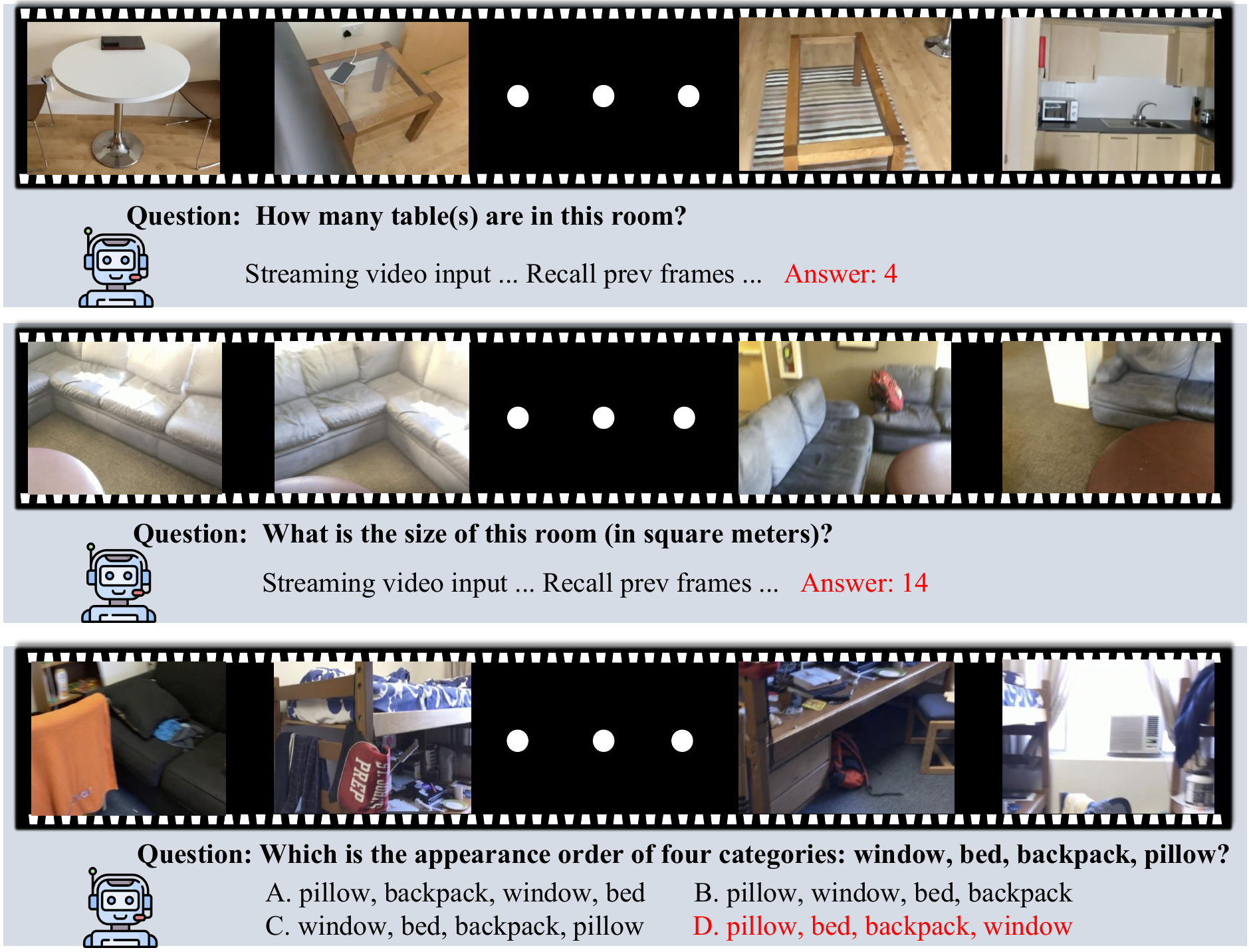}
    \caption{QA Example Visualization on VSI-Bench}
    \label{fig:visualization_VSI_Bench}
\end{figure}

\begin{figure}[!h]
    \centering
    \includegraphics[width=0.9\linewidth]{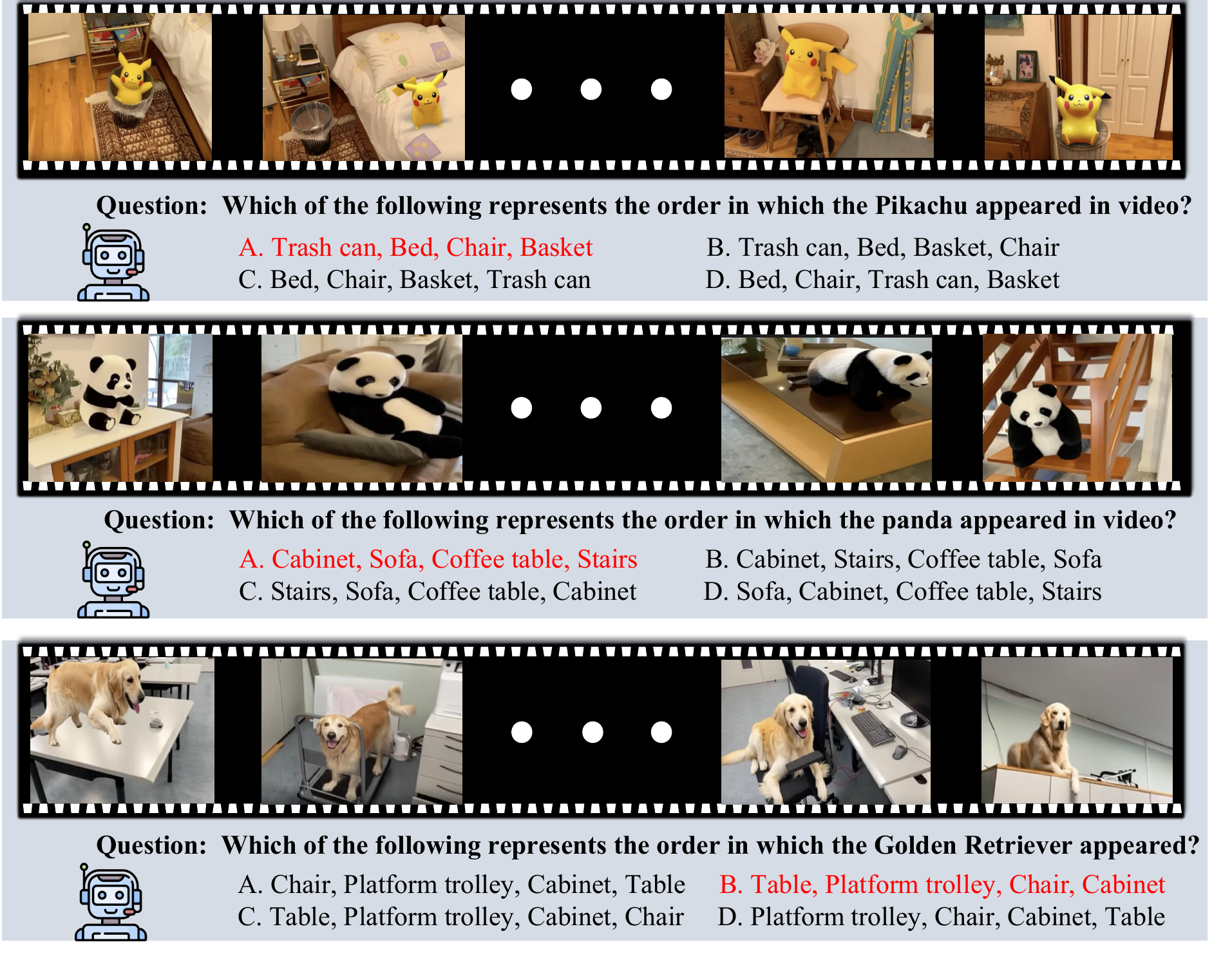}
    \caption{QA Example Visualization on VSI-SUPER-RECALL}
    \label{fig:visualization_VSR}
\end{figure}

\subsection{QA Examples Visualization}
We provide QA example visualizations on the four evaluation benchmarks. Fig.~\ref{fig:visualization_VSI_Bench} shows examples on VSI-Bench, Fig.~\ref{fig:visualization_VSR} on VSI-SUPER-RECALL, Fig.~\ref{fig:visualization_VSI_SUPER_COUNT} on VSI-SUPER-COUNT, and Fig.~\ref{fig:visualization_mindcube} on MindCube.

\begin{figure}[!t]
    \centering
    \includegraphics[width=0.9\linewidth]{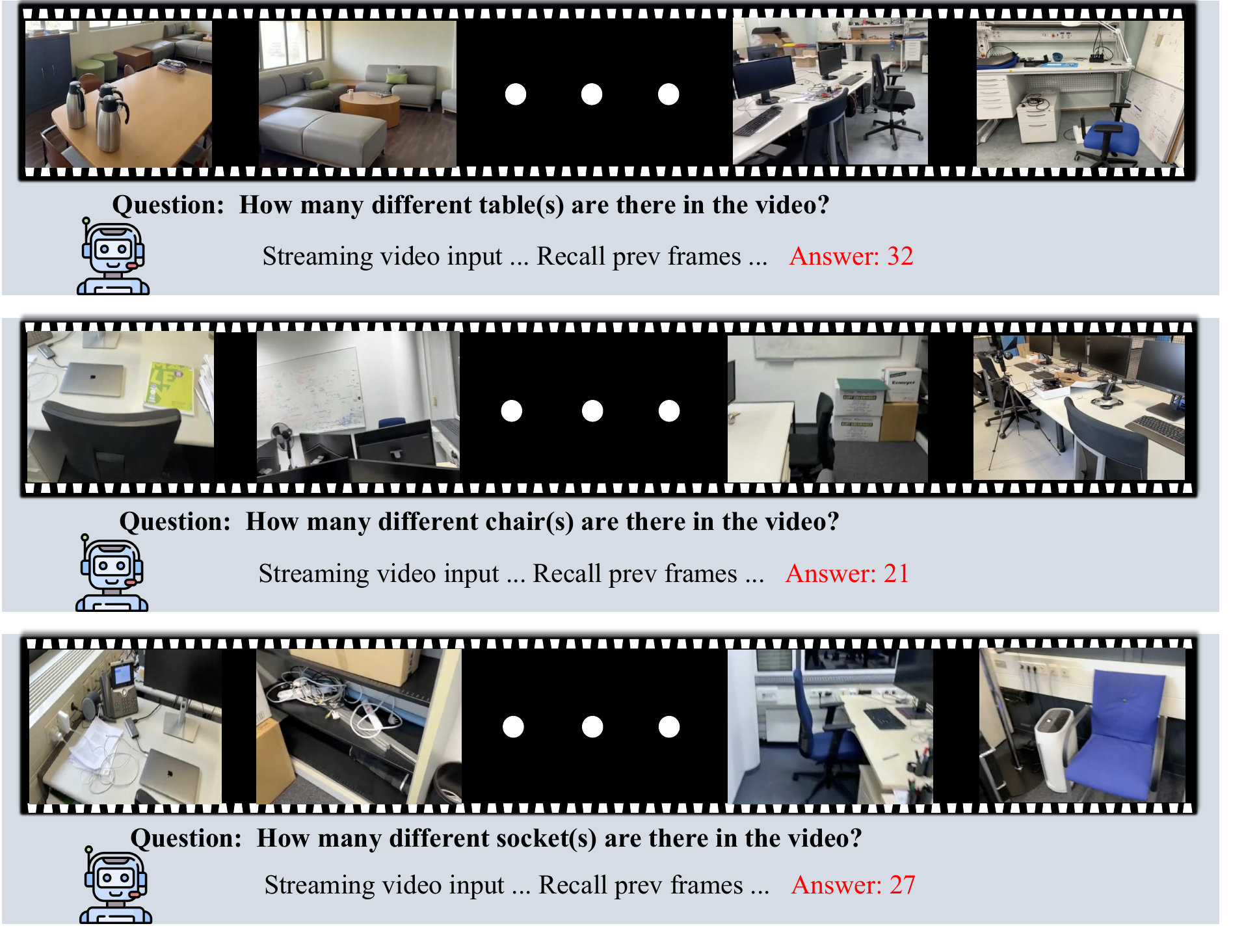}
    \caption{QA Example Visualization on VSI-SUPER-COUNT}
    \label{fig:visualization_VSI_SUPER_COUNT}
\end{figure}

\begin{figure}[!h]
    \centering
    \includegraphics[width=0.9\linewidth]{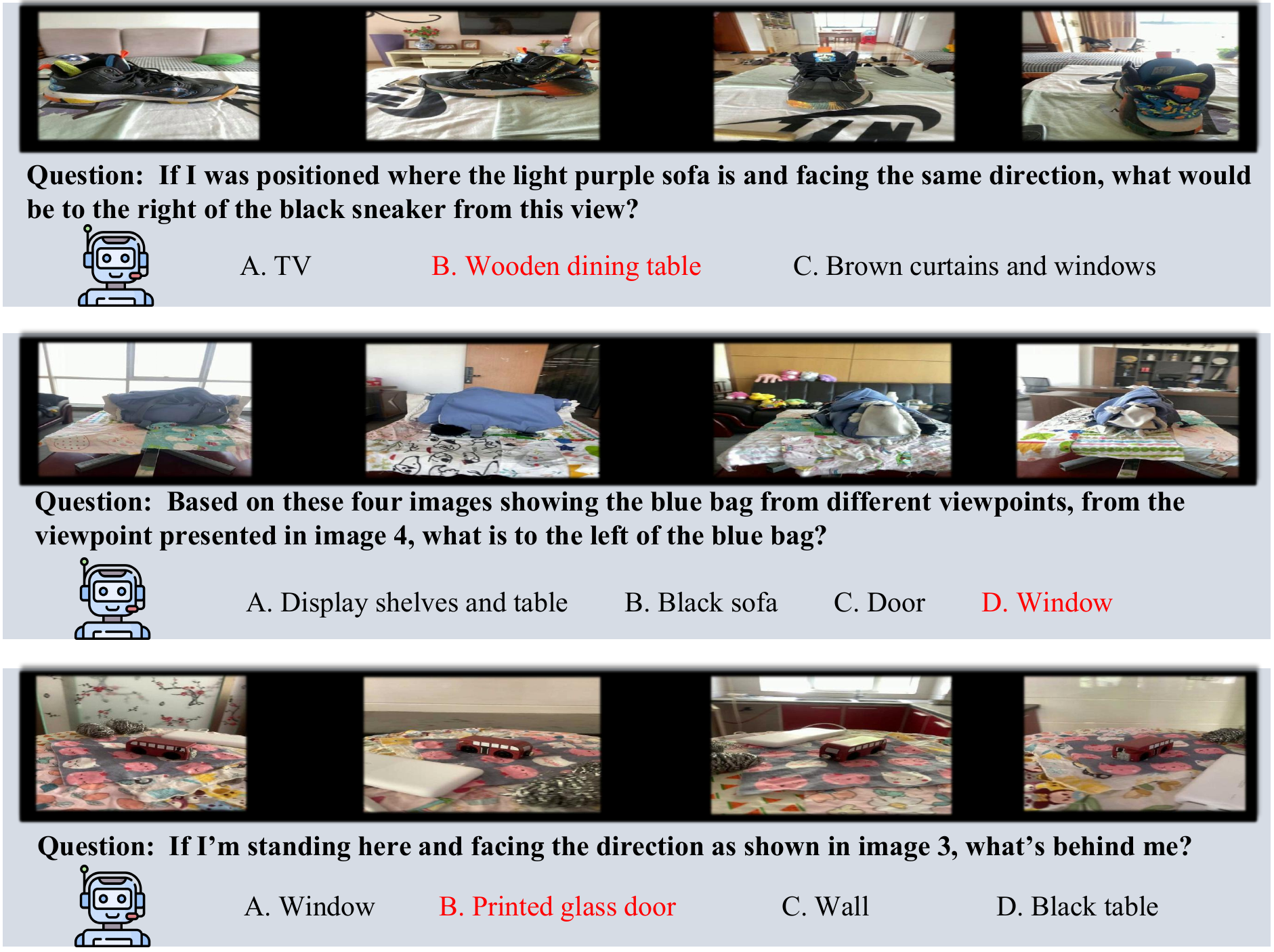}
    \caption{QA Example Visualization on MindCube}
    \label{fig:visualization_mindcube}
\end{figure}

\end{document}